\documentclass{article}

\usepackage[preprint,nonatbib]{neurips_2026}

\usepackage[utf8]{inputenc} 
\usepackage[T1]{fontenc}    
\usepackage{hyperref}       
\usepackage{url}            
\usepackage{microtype}      
\usepackage{xcolor}         
\usepackage{graphicx}
\usepackage{subcaption}
\usepackage{booktabs}       
\usepackage{amsfonts}       
\usepackage{nicefrac}       
\usepackage{makecell}
\usepackage{float}
\usepackage{placeins}

\usepackage{amsmath}
\usepackage{amssymb}
\usepackage{mathtools}
\usepackage{amsthm}
\usepackage{longtable}
\usepackage{multirow}
\usepackage{enumitem}
\usepackage[numbers,sort&compress]{natbib}

\usepackage[capitalize,noabbrev]{cleveref}

\theoremstyle{plain}

\theoremstyle{definition}

\theoremstyle{remark}

\usepackage[textsize=tiny]{todonotes}

\title{GeoTransolver: Learning Physics on Irregular Domains using Multi-scale Geometry Aware Physics Attention Transformer}

\author{
  Corey Adams\thanks{Equal contribution.},\;
  Rishikesh Ranade\footnotemark[1],\;
  Sheel Nidhan,\;
  Mohammad Amin Nabian \\
  \bfseries Deepak Akhare,\;
  Ram Cherukuri,\;
  Sanjay Choudhry \\
  \mdseries NVIDIA Corporation, Santa Clara, US \\
  \texttt{coreya@nvidia.com, rranade@nvidia.com}
}

\begin{document}

\maketitle

\begin{abstract}
    We present GeoTransolver, a multiscale geometry-aware physics
    attention transformer for Computer Aided Engineering (CAE).
    GeoTransolver extends the Transolver backbone with GALE
    (Geometry-Aware Latent Embeddings) attention, which pairs
    physics-aware self-attention on learned state slices with
    cross-attention to a shared geometry and global context computed
    via multi-scale ball queries (inspired by Domino) and reused in
    every block. Implemented and released in NVIDIA PhysicsNeMo,
    GeoTransolver persistently projects geometry and global parameters,
    into physical state spaces to anchor
    computations to domain structure and operating regimes. We
    benchmark on DrivAerML, SHIFT-SUV, and SHIFT-Wing against Domino,
    Transolver (PhysicsNeMo implementation), and literature-reported
    AB-UPT, evaluating drag/lift $R^2$ and relative $L_1$ errors on
    field variables. As an additional nonlinear structural mechanics 
    application, we also report Transolver and GeoTransolver results 
    on bumper-beam and full-vehicle Body-in-White (BIW) crash-dynamics 
    benchmarks, evaluating relative $L_2$ trajectory error and probe-level 
    kinematic MSE. GeoTransolver delivers improved accuracy,
    robustness to geometry and regime shifts, and favorable data
    efficiency; we include DrivAerML ablations and qualitative
    contour and design-trend results, advancing operator learning for
    high-fidelity surrogates on complex, irregular, non-linear
    domains.

\end{abstract}
    
\section{Introduction}
Computer-aided engineering (CAE) increasingly depends on AI-based surrogates 
to accelerate design exploration and reduce reliance on costly high-fidelity 
simulation \cite{zhao2025review, filippos2025advancing, brunton2020machine, 
sharma2023review}. Despite rapid progress in neural operators and 
physics-informed architectures, deploying these models on industrial CAE 
workloads exposes three persistent failure modes that limit their 
trustworthiness for engineering decision-making:

\begin{enumerate}[nosep]
    \item \textbf{Geometry dilution across depth.}
    Industrial meshes contain millions of surface and volume elements with 
    heterogeneous refinement near boundaries, wakes, and separation regions. 
    In transformer-based operators that inject geometric information only at 
    the input layer, boundary and shape detail is progressively diluted as 
    latent representations evolve through successive attention blocks. Later 
    layers lose access to fine-grained domain structure, degrading 
    predictions precisely where near-wall physics matters most -- for 
    example, wall-shear stress near the A-pillar or wheel arches of a 
    vehicle, or contact surfaces in crash dynamics 
    \cite{ashton2025fluid, wu2024transolver}.

    \item \textbf{Scale mismatch on non-uniform discretizations.}
    Automotive and aerospace meshes exhibit orders-of-magnitude
    differences in cell size between boundary-layer regions and far-field
    elements. A 
    single-radius neighborhood either floods with boundary-layer detail or 
    misses it entirely; global attention with uniform tokens averages over 
    the disparity. Capturing both near-boundary gradients and far-field 
    coupling requires explicit multi-scale geometric aggregation
    that spans both regimes simultaneously.

    \item \textbf{Single-shot regime conditioning.}
    CAE surrogates must generalize across operating regimes (Mach number,
    angle of attack, impact velocity) and boundary conditions. Existing
    conditional operator learners typically concatenate regime parameters
    to input tokens -- a single conditioning event that does not persist
    as the latent state evolves through depth. Just as U-Net skip
    connections were introduced to prevent high-resolution spatial features
    from being lost through successive pooling layers
    \cite{ronneberger2015u}, persistent conditioning is needed to prevent
    geometry and regime information from being washed out through
    successive attention layers. Without it, the model's internal state
    becomes decoupled from the imposed operating conditions in deeper
    layers.
\end{enumerate}

This paper introduces GeoTransolver, a transformer architecture for CAE
in which each design choice directly addresses one of these domain-specific
failure modes. GeoTransolver builds on the Transolver 
\cite{wu2024transolver} backbone and introduces Geometry-Aware Latent
Embeddings (GALE) attention, which pairs physics-aware self-attention on
learned state slices with cross-attention to a shared geometry and global
context at every block. To address scale mismatch, multi-scale ball queries
with radii spanning boundary-layer to far-field scales extract local
neighborhood features; geometry and global operating parameters are then
projected into the same latent slice-token space used by the attention
layers, providing persistent conditioning that prevents both geometry
dilution and regime drift. Prior methods condition on geometry only at the
input \cite{ranade2025domino, alkin2025ab} or at encoder/decoder
boundaries \cite{wen2025goat}; GeoTransolver maintains geometry and regime
context in the same learned basis as the evolving latent state at every
layer.

GeoTransolver is implemented and released as open-source software in the
NVIDIA PhysicsNeMo framework \cite{nvidia2025physicsnemo}, which provides
geometry data pipelines, multi-physics conditioning utilities, and scalable
training/inference infrastructure necessary for high-resolution CAE
workloads.

We benchmark GeoTransolver against Domino \cite{ranade2025domino}, AB-UPT
\cite{alkin2025ab}, and Transolver \cite{wu2024transolver} on three
industrial CAE datasets: DrivAerML \cite{ashton2024drivaerml} (500
parametric vehicle variants on 140M-element meshes), Luminary SHIFT-SUV
\cite{luminaryshiftsuv}, and Luminary SHIFT-Wing
\cite{luminaryshiftwing}, using the benchmarking framework in PhysicsNeMo
\cite{tangsali2025benchmarking}. These tasks span aerodynamic flows over
parameterized vehicle and wing geometries across diverse operating regimes.
We evaluate field-level reconstruction (velocity, pressure, wall shear) and
integral quantities (drag and lift coefficients), along with
generalization to unseen geometries and design trends. We further include
crash-dynamics benchmarks (bumper-beam impact and full-vehicle
Body-in-White frontal impact) to evaluate whether geometry-aware
conditioning transfers beyond external aerodynamics to transient,
contact-driven structural mechanics with large deformations, and to
demonstrate that the geometric context pathway can be paired with
alternative efficient attention backends
\cite{puri2025flare}.

Our contributions are two-fold:
\begin{enumerate}[nosep]
    \item \textbf{A geometry-persistent transformer design (GeoTransolver)}
    in which each architectural choice addresses a specific failure mode:
    GALE cross-attention at every block prevents geometry dilution
    (failure mode~1); multi-scale ball queries with radii spanning
    boundary-layer to far-field scales resolve the scale mismatch
    (failure mode~2); and a global-parameter context projector that
    injects operating-regime information at every depth prevents regime
    drift (failure mode~3). The geometry-aware context enters through a
    modular cross-attention path, so the self-attention backend can be
    swapped -- we demonstrate this by replacing physics attention with
    FLARE \cite{puri2025flare} on crash-dynamics
    benchmarks.

    \item \textbf{Industrial-scale evaluation and open-source release}
    within NVIDIA PhysicsNeMo, benchmarked on DrivAerML (500 vehicle
    variants, 140M-element meshes), SHIFT-SUV, SHIFT-Wing, and two
    crash-dynamics datasets. GeoTransolver delivers improved accuracy and
    robustness across all benchmarks, with ablations validating each
    design choice independently.
\end{enumerate}

\section{Background and related work}
AI-based surrogates for CAE span operator learning, mesh/geometry-aware 
representations, multi-scale architectures, and physics-informed training. 
Below we survey the most relevant lines of work and position GeoTransolver 
relative to them.

\subsection{Neural operators}
Neural operators learn mappings between function spaces for
resolution-agnostic PDE surrogates \cite{kovachki2023neural}.
DeepONet \cite{lu2021learning} introduced branch--trunk architectures;
FNO \cite{li2020fourier, li2020multipole} uses spectral convolution for
nonlocal interactions; and GNO \cite{li2020neural, li2023fourier} targets
irregular discretizations via message passing.
Transformer-based operators \cite{vaswani2017attention, hao2023gnot,
bryutkin2024hamlet, shih2025transformers, liu2025geometry, ovadia2024vito,
wu2024transolver} improve long-range coupling via global attention over
latent fields.
GeoTransolver replaces standard attention with GALE, structuring
self-attention around physics-aware state slices and cross-attending to
geometry and global context at every block.

\subsection{Geometry and mesh-aware encoders}

MeshGraphNet and related GNNs \cite{pfaff2020learning, pelissier2024graph, 
pilva2022learning, gladstone2024mesh} popularized message passing on simulation
meshes; extensions add learnable stencils and multigrid coarsening
\cite{nabian2024x, fortunato2022multiscale}. Point-cloud networks
\cite{choy2025factorized, chen2025tripnet, eliasof2021pde} introduced
multi-scale local aggregation for unstructured geometry.
Domino \cite{ranade2025domino} demonstrated that multi-scale ball queries
paired with global shape descriptors improve generalization across shape
families. AB-UPT \cite{alkin2025ab} explores unified physics transformers
with geometry-grounded tokens.

GAOT \cite{wen2025goat} is the most closely related prior work:
it combines multiscale attentional graph neural operator (GNO) encoders
and decoders with geometry embeddings, achieving strong results on
DrivAerML and other industrial-scale datasets. A concurrent 3D extension \cite{wen2025goat} further scales this approach. The key architectural
difference is \emph{where} geometry enters the network: GAOT injects
geometric information at encoder and decoder boundaries, so the
transformer processor layers between them operate without direct
geometric recall. GeoTransolver instead projects geometry into the same
slice-token basis used by the latent state and cross-attends at every
layer, providing persistent conditioning throughout the forward pass.
The two approaches also differ in metrics (GAOT reports MSE/MAE;
GeoTransolver reports relative $L_1$ and $R^2$), precluding direct
numerical comparison on shared datasets at this time.
Beyond geometry-aware encoders, U-Net hierarchies, graph coarsening,
and multigrid networks
\cite{ranade2022composable, liu2021multi, liu2023multiresolution}
and spectral methods \cite{li2020fourier, guibas2021adaptive} also
target multi-scale coupling, the latter struggling on irregular
domains. GeoTransolver addresses the scale-mismatch failure mode by
using multi-scale ball-query neighborhoods while preserving a
dedicated global-context channel via cross-attention at every depth.

\subsection{Physics-aware conditioning and constraints}

PINNs \cite{raissi2019physics} and PINO \cite{li2024physics} enforce PDEs
via residual losses; equivariant architectures
\cite{satorras2021n, brandstetter2021geometric, liao2022equiformer}
encode symmetries (e.g., SE(3)); and conditional operator learning
introduces regime metadata but typically conditions only once, risking
representation drift.
GeoTransolver projects geometry and global context into aligned
physical-state tokens at every block for persistent conditioning, and
is compatible with PINN/PINO-style losses.

\subsection{AI modeling for CAE}
Industrial CAE surrogates must handle irregular, heterogeneous
geometries (vehicles, wings, structural assemblies), capture multiscale
physics from boundary layers to far-field wakes, and remain conditioned
on operating regimes across design families. As identified in
Section~1, these requirements converge on three failure modes that
expose fundamental gaps in current architectures: geometry dilution
across depth, scale mismatch on non-uniform discretizations, and
single-shot regime conditioning. No single prior method addresses all
three.

Table~\ref{tab:related-positioning} maps each baseline to the three
failure modes. Methods that encode geometry only at the input
(Domino, AB-UPT) or at encoder/decoder boundaries (GAOT) leave later
transformer layers without direct geometric recall. Methods without
multi-scale neighborhoods (FNO, DeepONet, Transolver) cannot resolve
the 1000:1 resolution ratios typical of industrial meshes. No prior
method provides persistent context conditioning at every layer in the
same latent basis as the evolving physical state. GeoTransolver is
designed to close all three gaps simultaneously.


\begin{table}[t]
\centering
\caption{Failure-mode coverage across methods. Only GeoTransolver
provides persistent geometric context at every transformer layer.}
\label{tab:related-positioning}
\setlength{\tabcolsep}{3pt}
\begin{small}
\begin{tabular}{@{}l cccc cccc !{\vrule width 0pt} c@{}}
\toprule
& FNO & DeepONet & Transolver & \makecell{Mesh\\GraphNet}
  & DoMINO & AB-UPT & GAOT
  & \textbf{\makecell{Geo\\Transolver}} \\
\midrule
\makecell[l]{Geometry\\encoding}
  & \makecell{spectral\\grid}
  & \makecell{branch\\net}
  & \makecell{input\\tokens}
  & \makecell{edge\\features}
  & \makecell{ball\\queries}
  & \makecell{geo\\tokens}
  & \makecell{GNO\\(multi-sc.)}
  & \textbf{\makecell{ball queries\\+ projector}} \\
\addlinespace[2pt]
Multi-scale
  & ---
  & ---
  & ---
  & local
  & \makecell{multi-\\radii}
  & \checkmark
  & \checkmark
  & \textbf{\makecell{multi-\\radii}} \\
\addlinespace[2pt]
\makecell[l]{Context\\depth}
  & ---
  & ---
  & ---
  & ---
  & \makecell{input\\only}
  & \makecell{input\\only}
  & \makecell{enc/dec\\only}
  & \textbf{\makecell{every layer\\(x-Attn)}} \\
\bottomrule
\end{tabular}
\end{small}
\end{table}

\section{GeoTransolver}

GeoTransolver takes three inputs -- query points with local field
values, a geometry surface mesh, and optional global operating
parameters (e.g., Mach number, angle of attack) -- and predicts output
fields at the query positions. Each component of the architecture
addresses one of the three failure modes from Section~1: a shared
\emph{context} encodes geometry and global parameters in a latent
slice-token space and is reused at every layer to prevent geometry
dilution; multi-scale ball queries within the context construction
resolve boundary-layer-to-far-field scale mismatch and additionally
augment input tokens with local geometric features; and a global
parameter projector injects regime information into the same
slice-token space, providing persistent regime conditioning across
depth. Local field embeddings then pass through a stack of GALE
(Geometry-Aware Latent Embedding) blocks, each performing
physics-aware self-attention on learned slices followed by
cross-attention to the shared context, blended via a learnable gate
(Figure~\ref{fig:GALE}).

\paragraph{Notation.}
Throughout this section we use: $N$ for the number of query points,
$M$ for geometry points, $H$ for attention heads, $S$ for slices
(latent states) per head, $D$ for head dimension, $d_f$ for input
feature dimension, $d_p$ for global parameter dimension, $d_h$ for
geometric processor output dimension, and $N_s$ for the number of
ball-query scales.

\subsection{GeoTransolver inputs}

GeoTransolver operates on a set of $N$ query points, input Geometry
$\mathcal{G}$, and global operating parameters
$\mathbf{p}$:

\begin{enumerate}[nosep]
    \item \textbf{Query points} with positions and features:
    \begin{equation}
        X = \{(\mathbf{x}_{i},\, \mathbf{f}_{i})\}_{i=1}^{N},
        \quad \mathbf{x}_{i}\in\mathbb{R}^3,
        \quad \mathbf{f}_{i}\in\mathbb{R}^{d_f}
    \end{equation}
    where $\mathbf{f}_{i}$ may include surface normals, signed distance
    fields, Fourier embeddings, or any task-specific encoding. The model 
    predicts output fields at each query position $\mathbf{x}_i$.
    \item \textbf{Geometry point cloud}
    $\mathcal{G} = \{\mathbf{g}_j\}_{j=1}^{M}$,
    $\mathbf{g}_j \in \mathbb{R}^{3}$, representing the surface mesh
    or boundary of the domain.
    \item \textbf{Global parameters}
    $\mathbf{p} \in \mathbb{R}^{d_p}$ (e.g., Mach number, angle of
    attack, Reynolds number).
\end{enumerate}

Both the geometry $\mathcal{G}$ and global parameters $\mathbf{p}$ are
projected into the shared geometry context
as described in Section~\ref{sec:context}, which continuously augments the
model's latent space.

\subsection{Persistent geometry recall via GALE}

Transolver's physics attention \cite{wu2024transolver} avoids the
quadratic cost of full attention by projecting input tokens onto a small
set of learned \emph{slices} (latent basis vectors) and performing
self-attention in that reduced space, scaling to input sizes where full
self-attention is infeasible.

This design is efficient but introduces the \emph{geometry dilution}
failure mode identified in Section~1: because each layer receives only the
output of the previous layer, geometric and boundary information present
in the original inputs isz progressively diluted with depth. Later layers
have no mechanism to recall fine-grained domain structure, degrading
predictions in near-wall regions where boundary-layer physics depends on
local geometry. GALE (Geometry-Aware Latent Embeddings) addresses this
directly by adding cross-attention from the slice tokens to a shared
context that encodes geometry and global parameters in the same
slice-token space, ensuring every layer retains direct access to domain
structure.

\begin{figure}
    \centering
    \includegraphics[width=0.9\linewidth]{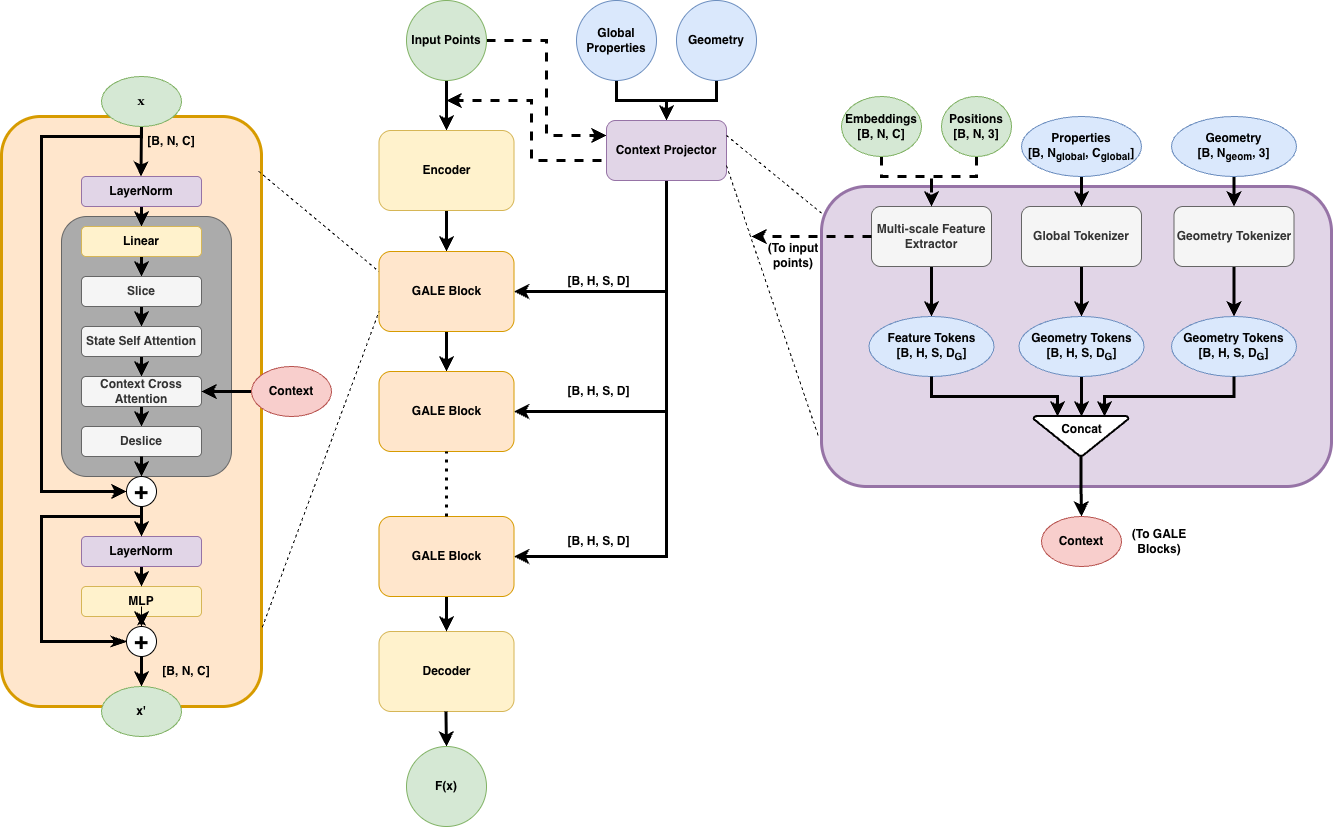}

    \caption{GeoTransolver architecture. \textbf{Right:} the context
    projector constructs a shared geometry and global context by projecting
    multi-scale ball-query features and global parameters into slice-token
    space via a half-attention operation; learned geometric features are
    also concatenated onto local input tokens. \textbf{Center:} The 
    learned context is continually injected to the 
    model to provide persistent geometrical recall. 
    \textbf{Left:} each GALE block extends Transolver's Physics Attention with
    the addition of cross attention to the shared global context.}
    
    \label{fig:GALE}
\end{figure}

The key deviations for a GALE layer occur after the latent state self
attention, as seen in Figure \ref{fig:GALE} (left).  The learned projections
for a given layer are used in a cross-attention mechanism with the `context'
embedding, learned directly from the input points and geometries. 
This cross-attention is then blended with the self-attention output
via a learnable scalar gate $\alpha = \sigma(\eta)$, where $\eta$ is
a per-layer parameter and $\sigma$ is the sigmoid function. This allows
the model to adapt its balance between geometric context and
self-attended state at each depth. After mixing, the slice tokens are
de-sliced back to the input dimensionality as in Transolver
\cite{wu2024transolver}.

Although we use Transolver's physics attention as the default self-attention backend, the GALE formulation is not restricted to this choice. The geometry-aware context enters through a separate cross-attention path, so the self-attention branch can be replaced by other efficient token-mixing mechanisms. In the crash-dynamics experiments, we include a lightweight backend-swap demonstration in which the physics-attention backend is replaced with FLARE attention \cite{puri2025flare} while keeping the GeoTransolver geometry-context pathway unchanged.

\subsection{Multi-scale context for boundary-layer to far-field coupling}
\label{sec:context}

The second and third failure modes -- scale mismatch and single-shot
regime conditioning -- are addressed by the context construction stage.
Before any GALE block is applied, GeoTransolver builds a shared
\emph{context} that encodes the geometry $\mathcal{G}$ and global
parameters $\mathbf{p}$ in the same slice-token space used by GALE
attention. The context is assembled from two reusable building blocks --
a \emph{context projector} $\mathcal{T}$ and a \emph{geometric processor}
$\Pi$ -- composed in three independent paths: a geometry tokenizer,
a global tokenizer, and a multi-scale feature extractor
(see Figure~\ref{fig:GALE}, right).
Each path produces slice tokens in the same latent space as the GALE
attention slices; the outputs are concatenated into a single context
tensor $C$, computed once per input and reused at every depth.

The multi-scale extractor also produces geometric embedding for each input query 
$\mathbf{x}_i$
that are concatenated onto each input token before the first GALE
block, giving every point direct access to nearby boundary geometry information.

\subsubsection{Context projector}
The context projector $\mathcal{T}$ implements the encoding half of
Transolver's physics attention~\cite{wu2024transolver}: it maps a
variable-length token sequence into a fixed set of \emph{slice tokens}
in the same latent space as the GALE attention slices, but never
decodes back.
Given $N$ tokens $\mathbf{X}\in\mathbb{R}^{N\times C}$, learned
projections yield $\mathbf{Q},\mathbf{K}\in\mathbb{R}^{H\times N\times D}$.
A per-head map $W_s\in\mathbb{R}^{D\times S}$ produces slice logits
$\mathbf{P}=\mathbf{Q}W_s$; temperature-scaled softmax and normalised
aggregation with $\mathbf{K}$ yield slice tokens $\mathbf{Z}$:

\begin{equation}\label{eq:context-proj}
    \mathbf{A} = \mathrm{softmax}_S\!\big(\mathbf{P}/\tau\big),
    \qquad
    \mathbf{Z} = \overline{\mathbf{A}}^{\!\top}\mathbf{K}
      \;\in\;\mathbb{R}^{H\times S\times D},
\end{equation}

where $\overline{\mathbf{A}}$ denotes column-normalised $\mathbf{A}$.
The same architecture $\mathcal{T}$ is instantiated with separate
learned parameters for each context source below.

\subsubsection{Geometric processor}

A \emph{geometric processor} $\Pi_{r_s,k_s}(\mathbf{q},\,\mathbf{t})$
extracts local geometric context at scale $s$.
For each point in $\mathbf{q}\in\mathbb{R}^{N_q\times 3}$, a ball
query retrieves up to $k_s$ neighbours within radius $r_s$ from
$\mathbf{t}\in\mathbb{R}^{N_t\times 3}$; the flattened coordinates
are mapped by a learned MLP $\phi_s$ to a bounded embedding:

\begin{align}\label{eq:geo-proc}
    \mathcal{N}_{r_s,k_s}(\mathbf{q},\,\mathbf{t})
      &= \mathrm{BQ}_{r_s,k_s}(\mathbf{q},\,\mathbf{t})
      \;\in\;\mathbb{R}^{N_q \times k_s \times 3},
      \\
    \mathbf{v}_{s}
      &= \mathrm{flatten}\!\big(
            \mathcal{N}_{r_s,k_s}(\mathbf{q},\,\mathbf{t})
         \big)
      \;\in\;\mathbb{R}^{N_q \times 3k_s},
      \\
    \Pi_{r_s,k_s}(\mathbf{q},\,\mathbf{t})
      &= \tanh\!\big(\phi_s(\mathbf{v}_{s})\big)
      \;\in\;\mathbb{R}^{N_q \times d_h}.
\end{align}

Each geometric processor is parameter-light and independently
parameterised; one is instantiated per scale and per cross-cloud
direction.

\subsubsection{Context construction}
The three context paths below each produce slice tokens via the
context projector (Eq.~\ref{eq:context-proj}); the results are
concatenated along the feature dimension to form the shared context
$C$.

\paragraph{Geometry and global tokenizers.}
The geometry point cloud
$\mathcal{G}=\{\mathbf{g}_j\}_{j=1}^{M}$ and the global operating
parameters $\mathbf{p}\in\mathbb{R}^{d_p}$ (e.g.\ Mach number, free-stream velocity)
are each projected into slice-token space by independent
context projectors:
\begin{equation}\label{eq:geo-tok}
    \mathbf{Z}_{\mathrm{geo}}
      = \mathcal{T}_{\mathrm{geo}}\!\big(\mathcal{G}\big),
    \qquad
    \mathbf{Z}_{\mathrm{glob}}
      = \mathcal{T}_{\mathrm{glob}}\!\big(\mathbf{p}\big)
      \;\in\;\mathbb{R}^{H\times S\times D}.
\end{equation}
No de-slicing is performed; both outputs remain in slice-token space
for cross-attention context at every GALE layer.

\paragraph{Multi-scale feature extractor.}

The multi-scale feature extractor applies geometric processors
$\Pi_{r_s,k_s}$ at $N_s$ scales $\{(r_s,k_s)\}_{s=1}^{N_s}$, where
short-range radii capture boundary-layer detail and long-range radii
enable non-local coupling (Section~\ref{ablation}).
Each scale uses $\Pi_{r_s,k_s}$ in two symmetric cross-cloud
directions and a per-scale context projector $\mathcal{T}_s$.

\emph{Input-to-geometry (context path).}
For each geometry point $\mathbf{g}_j$, nearby input points $\mathbf{x}$ are gathered, processed,
and tokenized into slice space:
\begin{align}\label{eq:ctx-path}
    \mathbf{h}^{\mathrm{ctx}}_s
      &= \Pi_{r_s,k_s}\!\big(\mathbf{x},\;\mathbf{g}\big)
      \;\in\;\mathbb{R}^{M\times d_h},
      \\
    \mathbf{Z}_s
      &= \mathcal{T}_s\!\big(\mathbf{h}^{\mathrm{ctx}}_s\big)
      \;\in\;\mathbb{R}^{H\times S\times D}.
\end{align}

\emph{Geometry-to-input (local augmentation path).}
Symmetrically, for each input point $\mathbf{x}_i$, nearby geometry
points are gathered:
\begin{equation}\label{eq:aug-path}
    \mathbf{h}^{\mathrm{loc}}_s
      = \Pi_{r_s,k_s}\!\big(\mathbf{g},\;\mathbf{x}\big)
      \;\in\;\mathbb{R}^{N\times d_h}.
\end{equation}

Per-scale features are concatenated to form a local augmentation
vector
$\mathbf{u} = [\mathbf{h}^{\mathrm{loc}}_1,\dots,
\mathbf{h}^{\mathrm{loc}}_{N_s}]
\;\in\;\mathbb{R}^{N\times N_s d_h}$,
which is appended to each input token before the first GALE block,
giving every query point direct access to nearby boundary geometry.

\paragraph{Assembled context.}
The shared context $C$ concatenates all slice tokens along the feature
dimension:
\begin{equation}\label{eq:context}
    C = \bigl[\,\mathbf{Z}_{\mathrm{geo}},\;\;
         \mathbf{Z}_{\mathrm{glob}},\;\;
         \mathbf{Z}_1,\,\dots,\,\mathbf{Z}_{N_s}\,\bigr]
      \;\in\;\mathbb{R}^{H\times S\times d_c},
\end{equation}
where $d_c = (2 + N_s)\cdot D$ and the concatenation is along the
last (feature) dimension.  All three sources share the same slice
count $S$ and head count $H$.

Separately, the local augmentation vectors
$\mathbf{u}_i\in\mathbb{R}^{N_s d_h}$ from the geometry-to-input path
(Eq.~\ref{eq:aug-path}) are concatenated onto each input token after
the initial MLP projection and before the first GALE block:
\begin{equation}\label{eq:input-aug}
    \tilde{\mathbf{x}}_i
      = \bigl[\,\mathrm{MLP}(\mathbf{f}_i),\;\;\mathbf{u}_i\,\bigr]
      \;\in\;\mathbb{R}^{d_{\mathrm{model}} + N_s d_h}.
\end{equation}
Together, $C$ provides persistent geometric and global context via
cross-attention at every GALE layer, while $\mathbf{u}_i$ enriches the
initial per-point representation with local boundary information.
Both are computed once per forward pass.

\section{Experiment details}

\subsection{Datasets}

The benchmarking, validation and ablation studies are carried out on 3 datasets, DrivAerML \cite{ashton2024drivaerml}, Luminary SHIFT-SUV \cite{luminaryshiftsuv} and Luminary SHIFT-Wing \cite{luminaryshiftwing}. As an additional example beyond the three aerodynamics datasets, we evaluate GeoTransolver on two crash-dynamics benchmarks: a bumper-beam impact dataset and a full-vehicle Body-in-White (BIW) frontal-impact dataset. Additional details about each of these datasets with information such as training, validation and testing splits are provided in the Appendix.

\subsection{Models}

We evaluate GeoTransolver on SHIFT‑SUV and SHIFT‑Wing against Transolver \cite{wu2024transolver}, DoMINO \cite{ranade2025domino}, and literature‑reported AB‑UPT \cite{alkin2025app}, and probe architectural sensitivity via an ablation on DrivAerML (Section \ref{ablation}). Additional configuration and training details are provided in the appendix. For the crash-dynamics benchmarks, we compare Transolver, GeoTransolver, and GeoTransolver with a FLARE attention \cite{puri2025flare} backend under matched training protocols.

\subsection{Metrics}
We evaluate field-level accuracy via the relative $L_1$ norm
${\sum_j\|u_j - \tilde{u}_j\|}\big/{\sum_j\|u_j\|}$,
and for surface predictions report drag and lift $R^2$ computed from
integrated surface forces
${F} = \oint_S (-(p_s - p_{\infty}){\hat{n}} + {\tau}_w)\,dS$.
For crash dynamics we report relative $L_2$ error over the
spatiotemporal structural response and probe-level kinematic MSE
(details in Appendix).

\section{Results} \label{results}

\subsection{DrivAerML and ablation studies} \label{ablation}

Table~\ref{tab:ablation-combined} reports relative $L_1$ errors (\%)
on the DrivAerML test set (48 designs) and consolidates ablations over
three architectural axes: GALE depth, multi-scale ball-query radii,
and ball-query kernel size. Each group varies one axis while holding
others at the best-found setting; bold marks the overall best row,
which achieves below-5\% surface and volume field error and
near-perfect drag/lift agreement.

\begin{table}[t]
\centering
\caption{Relative $L_1$ errors (\%) on DrivAerML test geometries across three ablation axes. $C_D$/$C_L$ are $R^2$ values.}
\label{tab:ablation-combined}
\begin{small}
\begin{tabular}{@{}ll l cccc cc@{}}
\toprule
& & & \multicolumn{4}{c}{Surface} & \multicolumn{2}{c}{Volume} \\
\cmidrule(lr){4-7} \cmidrule(lr){8-9}
Axis & Setting & Size & $p_s$ & $\tau_w$ & $C_D$ & $C_L$ & $p_v$ & $u$ \\
\midrule
\multirow{5}{*}{\rotatebox{90}{Layers}}
 & 6  & 9M  & 3.52 & 5.88 & 0.996 & 0.987 & 3.79 & 4.44 \\
 & 10 & 14M & 3.25 & 5.51 & 0.995 & 0.987 & 3.31 & 4.21 \\
 & 14 & 20M & 3.11 & 5.29 & 0.995 & 0.983 & 3.34 & 4.24 \\
 & 18 & 26M & 2.95 & 5.19 & 0.994 & 0.987 & 3.08 & 4.08 \\
 & 20 & 29M & \textbf{2.86} & \textbf{4.9} & \textbf{0.996} & \textbf{0.991} & \textbf{3.09} & \textbf{4.02} \\
\midrule
\multirow{4}{*}{\rotatebox{90}{Radii}}
 & $0.05$           & 12M & 3.14 & 5.38 & 0.993 & 0.989 & 3.60 & 4.34 \\
 & $2.5$            & 12M & 3.09 & 5.38 & 0.995 & 0.986 & 3.24 & 4.20 \\
 & $0.05$--$2.5$ (4) & 21M & 3.03 & 5.23 & 0.993 & 0.989 & 3.06 & 4.06 \\
 & $0.01$--$5.0$ (6) & 29M & \textbf{2.86} & \textbf{4.9} & \textbf{0.996} & \textbf{0.991} & \textbf{3.09} & \textbf{4.02} \\
\midrule
\multirow{3}{*}{\rotatebox{90}{Kernel}}
 & $k{=}8$  & 13M & 3.12 & 5.32 & 0.994 & 0.984 & 3.37 & 4.06 \\
 & $k{=}16$ & 18M & 3.07 & 5.27 & 0.993 & 0.989 & 3.12 & 4.11 \\
 & $k{=}32$ & 29M & \textbf{2.86} & \textbf{4.9} & \textbf{0.996} & \textbf{0.991} & \textbf{3.01} & \textbf{4.02} \\
\bottomrule
\end{tabular}
\end{small}
\end{table}

Across all three axes, accuracy improves monotonically with capacity,
with diminishing returns. The best configuration (20 layers, 6 radii
spanning $r{=}0.01$--$5.0$, $k{=}32$; 29M params) achieves the bold-row
results. Fine radii aid near-wall detail while large radii capture
global couplings; larger kernels enrich local geometric aggregation.
A separate study on query and geometry token counts
(Appendix Table~\ref{tab:results-l1-drivaerml-sampling}) shows that
adequate geometric coverage is critical for volume accuracy.

\paragraph{Computational cost.}
On 8$\times$H100 with batch size~8, single-scale ball queries add
1.1--1.2$\times$ throughput overhead and the 4- and 6-scale multi-radii
configurations add approximately 2$\times$;
the 6-scale config requires reducing query points from 200k to 160k to
fit within H100's 80~GB. Full per-configuration training and inference
throughput data are in
Appendix~\ref{sec:compute-cost-appendix}, Table~\ref{tab:compute-cost}.

\subsection{SHIFT-SUV, SHIFT-Wing, and crash dynamics}

Table~\ref{tab:shift-volume} reports relative $L_1$ errors on the
volume fields (pressure $p_v$ and velocity $u$) across all four
SHIFT test conditions: SHIFT-SUV Estate and Fastback bodies, and
SHIFT-Wing at Mach 0.5 and 0.85. GeoTransolver attains the lowest
error on every column except Estate $p_v$, where AB-UPT is lower
by $0.0001$. On surface
metrics (Appendix Table~\ref{tab:shift-surface}), GeoTransolver leads
on most fields with two exceptions: surface pressure at Mach 0.85
(AB-UPT) and wall-shear stress at Mach 0.5 (DoMINO, trained
per-Mach). For SHIFT-SUV drag and lift $R^2$, GeoTransolver achieves
the highest scores (Estate $C_D/C_L{=}0.98/0.88$, Fastback
$0.99/0.99$); on SHIFT-Wing all models reach $R^2{=}1.0$ on $C_D/C_L$
at both Mach numbers, so integral force metrics do not discriminate.

On the crash-dynamics benchmarks (Table~\ref{tab:crash_results}),
GeoTransolver reduces relative $L_2$ error vs.\ Transolver on both
bumper-beam and BIW frontal-impact, and replacing the
physics-attention backend with FLARE further improves both,
demonstrating that the geometry-aware context pathway is
backend-agnostic. Contours, design trends, deformation evolution, and
probe-level kinematic MSE are reported in the Appendix.

\begin{table}[t]
\centering
\caption{Volume-field relative $L_1$ errors (\%) across SHIFT-SUV
(Estate, Fastback) and SHIFT-Wing (Ma=0.5, Ma=0.85). DoMINO is
trained per-Mach on SHIFT-Wing; other models share the combined
dataset. Surface-field counterparts are in
Appendix Table~\ref{tab:shift-surface}.}
\label{tab:shift-volume}
\setlength{\tabcolsep}{4pt}
\begin{small}
\begin{tabular}{@{}l cc cc cc cc@{}}
\toprule
& \multicolumn{2}{c}{Estate} & \multicolumn{2}{c}{Fastback}
& \multicolumn{2}{c}{Ma=0.5} & \multicolumn{2}{c}{Ma=0.85} \\
\cmidrule(lr){2-3}\cmidrule(lr){4-5}\cmidrule(lr){6-7}\cmidrule(lr){8-9}
Model & $p_v$ & $u$ & $p_v$ & $u$ & $p_v$ & $u$ & $p_v$ & $u$ \\
\midrule
GeoTransolver & 0.0026 & \textbf{1.36} & \textbf{0.0023} & \textbf{1.30}
              & \textbf{0.022} & \textbf{1.92} & \textbf{0.099} & \textbf{2.00} \\
AB-UPT        & \textbf{0.0025} & 2.25 & 0.0024 & 2.21
              & 0.027 & 9.56 & 0.125 & 9.51 \\
DoMINO        & 0.0062 & 8.14 & 0.0067 & 7.73
              & 2.25  & 21.34 & 3.17 & 29.2 \\
Transolver    & 0.004  & 1.87 & 0.0039 & 1.81
              & 1.19  & 3.60 & 1.34 & 3.72 \\
\bottomrule
\end{tabular}
\end{small}
\end{table}

\begin{table}[t]
\centering
\caption{Crash-dynamics relative $L_2$ error on bumper-beam and BIW
frontal-impact datasets (lower is better).}
\label{tab:crash_results}
\begin{small}
\begin{tabular}{@{}l cc@{}}
\toprule
Model & Bumper Beam & BIW Crash \\
\midrule
Transolver & $9.12 \times 10^{-3}$ & $1.60 \times 10^{-2}$ \\
GeoTransolver & $7.32 \times 10^{-3}$ & $1.33 \times 10^{-2}$ \\
GeoTransolver with FLARE & $\mathbf{6.80 \times 10^{-3}}$ & $\mathbf{8.95 \times 10^{-3}}$ \\
\bottomrule
\end{tabular}
\end{small}
\end{table}

\section{Conclusion}
Industrial CAE surrogate modeling demands that transformer-based operators
retain geometry awareness across depth, resolve multi-scale features on
non-uniform meshes, and maintain regime conditioning throughout the
forward pass. In this work, we identified these three failure modes in
existing architectures and introduced GeoTransolver, a transformer design
in which each component directly addresses one: GALE cross-attention
prevents geometry dilution, multi-scale ball queries resolve
boundary-layer-to-far-field scale mismatch, and a global-parameter
context projector prevents regime drift.

We benchmarked GeoTransolver on DrivAerML, SHIFT-SUV, SHIFT-Wing, and
two crash-dynamics datasets, demonstrating improved accuracy and
robustness compared to Domino, Transolver, and AB-UPT. Ablations on
DrivAerML validate each design choice independently, confirming that
multi-scale radii, deeper GALE stacks, and adequate geometric token
coverage each contribute measurably. The crash-dynamics results further
show that geometry-aware conditioning transfers beyond aerodynamics to
transient, contact-driven structural mechanics, and that the GALE context
pathway is compatible with alternative attention backends (FLARE).
GeoTransolver is released as open-source software in NVIDIA PhysicsNeMo,
making it accessible to CFD researchers and application engineers.

\subsection{Limitations}
GeoTransolver has several limitations that frame the scope of the
results presented here. The multi-scale ball-query context pathway
increases per-step training and inference cost relative to
baseline (Table~\ref{tab:compute-cost}, $\sim$2.1$\times$).
The DrivAerML out-of-distribution designs (lowest
and highest drag) are never seen during training and remain the
hardest cases (see Appendix~A.1.1); accuracy on these
extremes can likely be improved with augmented training data but
is not directly addressed here. Finally, the crash-dynamics evaluation is
limited to two datasets (bumper beam and a single BIW frontal-impact
configuration) over short impact horizons, and broader transferability
of the geometry-aware context is an active research area.

\subsection{Broader impacts}
GeoTransolver targets industrial CAE surrogate modelling. On the
positive side, faster surrogate-based design iteration can reduce the
number of high-fidelity CFD/FEM runs required during product
development, lowering the HPC energy and carbon footprint of
engineering workflows, and the open-source release in NVIDIA
PhysicsNeMo lowers the barrier to high-fidelity CAE for academic
and industrial users. On the negative side, learned
surrogates that look qualitatively correct can be misused in
safety-critical decisions -- e.g., crashworthiness or structural
sign-off -- if their predictions are treated as ground truth without
validation against high-fidelity simulation or experiment.

{\small
\bibliography{references}

@article{zhao2025review,
  title={Review of empowering computer-aided engineering with artificial intelligence},
  author={Zhao, Xu-Wen and Tong, Xiao-Meng and Ning, Fang-Wei and Cai, Mao-Lin and Han, Fei and Li, Hong-Guang},
  journal={Advances in Manufacturing},
  pages={1--41},
  year={2025},
  publisher={Springer}
}

@article{filippos2025advancing,
  title={Advancing Fluid Mechanics with Artificial Intelligence and Machine Learning},
  author={Filippos, Sofos},
  journal={Fluids},
  volume={10},
  number={11},
  pages={297},
  year={2025},
  publisher={MDPI AG}
}

@article{brunton2020machine,
  title={Machine learning for fluid mechanics},
  author={Brunton, Steven L and Noack, Bernd R and Koumoutsakos, Petros},
  journal={Annual review of fluid mechanics},
  volume={52},
  number={1},
  pages={477--508},
  year={2020},
  publisher={Annual Reviews}
}

@article{sharma2023review,
  title={A review of physics-informed machine learning in fluid mechanics},
  author={Sharma, Pushan and Chung, Wai Tong and Akoush, Bassem and Ihme, Matthias},
  journal={Energies},
  volume={16},
  number={5},
  pages={2343},
  year={2023},
  publisher={MDPI}
}

@article{ashton2025fluid,
  title={Fluid Intelligence: A Forward Look on AI Foundation Models in Computational Fluid Dynamics},
  author={Ashton, Neil and Brandstetter, Johannes and Mishra, Siddhartha},
  journal={arXiv preprint arXiv:2511.20455},
  year={2025}
}

@article{vaswani2017attention,
  title={Attention is all you need},
  author={Vaswani, Ashish and Shazeer, Noam and Parmar, Niki and Uszkoreit, Jakob and Jones, Llion and Gomez, Aidan N and Kaiser, {\L}ukasz and Polosukhin, Illia},
  journal={Advances in neural information processing systems},
  volume={30},
  year={2017}
}

@article{luminaryshiftsuv,
  title={Shift-suv: High-fidelity computational fluid dynamics dataset for suv external aerodynamics},
  author={Luminary Cloud},
  journal={URL https://huggingface.co/datasets/luminary-shift/SUV/},
  year={2025}
}

@article{luminaryshiftwing,
  title={Shift-wing: High-fidelity computational fluid dynamics dataset for transonic wing external aerodynamics},
  author={Luminary Cloud},
  journal={URL https://huggingface.co/datasets/luminary-shift/WING/},
  year={2025}
}

@article{ashton2024drivaerml,
  title={DrivAerML: High-fidelity computational fluid dynamics dataset for road-car external aerodynamics},
  author={Ashton, Neil and Mockett, Charles and Fuchs, Marian and Fliessbach, Louis and Hetmann, Hendrik and Knacke, Thilo and Schonwald, Norbert and Skaperdas, Vangelis and Fotiadis, Grigoris and Walle, Astrid and others},
  journal={arXiv preprint arXiv:2408.11969},
  year={2024}
}

@article{tangsali2025benchmarking,
  title={A benchmarking framework for ai models in automotive aerodynamics},
  author={Tangsali, Kaustubh and Ranade, Rishikesh and Nabian, Mohammad Amin and Kamenev, Alexey and Sharpe, Peter and Ashton, Neil and Cherukuri, Ram and Choudhry, Sanjay},
  journal={arXiv preprint arXiv:2507.10747},
  year={2025}
}

@article{spalart2006new,
  title={A new version of detached-eddy simulation, resistant to ambiguous grid densities},
  author={Spalart, Philippe R and Deck, Shur and Shur, Michael L and Squires, Kyle D and Strelets, M Kh and Travin, Andrei},
  journal={Theoretical and computational fluid dynamics},
  volume={20},
  number={3},
  pages={181--195},
  year={2006},
  publisher={Springer}
}

@article{chaouat2017state,
  title={The state of the art of hybrid RANS/LES modeling for the simulation of turbulent flows},
  author={Chaouat, Bruno},
  journal={Flow, turbulence and combustion},
  volume={99},
  number={2},
  pages={279--327},
  year={2017},
  publisher={Springer}
}

@article{heinz2020review,
  title={A review of hybrid RANS-LES methods for turbulent flows: Concepts and applications},
  author={Heinz, Stefan},
  journal={Progress in Aerospace Sciences},
  volume={114},
  pages={100597},
  year={2020},
  publisher={Elsevier}
}

@inproceedings{ashton2022hlpw,
  title={HLPW-4/GMGW-3: Hybrid RANS/LES Technology Focus Group Workshop Summary},
  author={Ashton, Neil and Batten, Paul and Cary, Andrew W and Holst, Kevin R and Skaperdas, Vangelis},
  booktitle={AIAA Aviation 2022 Forum},
  pages={3293},
  year={2022}
}

@techreport{hupertz2022towards,
  title={Towards a standardized assessment of automotive aerodynamic CFD prediction capability-AutoCFD 2: Ford drivaer test case summary},
  author={Hupertz, Burkhard and Lewington, Neil and Mockett, Charles and Ashton, Neil and Duan, Lian},
  year={2022},
  institution={SAE Technical Paper}
}

@techreport{zhang2019introduction,
  title={Introduction of the AeroSUV-a new generic SUV model for aerodynamic research},
  author={Zhang, Chenyi and Tanneberger, Max and Kuthada, Timo and Wittmeier, Felix and Wiedemann, Jochen and Nies, Juliane},
  year={2019},
  institution={SAE Technical Paper}
}

@inproceedings{heft2012experimental,
  title={Experimental and numerical investigation of the DrivAer model},
  author={Heft, Angelina I and Indinger, Thomas and Adams, Nikolaus A},
  booktitle={Fluids engineering division summer meeting},
  volume={44755},
  pages={41--51},
  year={2012},
  organization={American Society of Mechanical Engineers}
}

@book{zhang2021aerodynamic,
  title={Aerodynamic study on the vehicle shape parameters with respect to ground simulation},
  author={Zhang, Chenyi},
  year={2021},
  publisher={Springer}
}

@book{economon2024,
  title={Luminary contribution to the autocfd4 workshop},
  author={T Economon D Engwirda J Ho M Mara and R Mansharamani},
  year={2024},
  publisher={URL https://autocfd.org/autocfd4/}
}

@inproceedings{krakos2025gpu,
  title={GPU-based and Adaptive Solution Technology for the 5th AIAA High Lift Prediction Workshop},
  author={Krakos, Joshua and Gomes, Pedro and Ho, Jonathan and Mara, Michael and S{\'a}ez, Gonzalo and Loseille, Adrien and Economon, Thomas and Alonso, Juan J},
  booktitle={AIAA SCITECH 2025 Forum},
  pages={0496},
  year={2025}
}

@article{alkin2025app,
  title={AB-UPT for Automotive and Aerospace Applications},
  author={Alkin, Benedikt and Kurle, Richard and Serrano, Louis and Just, Dennis and Brandstetter, Johannes},
  journal={arXiv preprint arXiv:2510.15808},
  year={2025}
}

@article{wu2024transolver,
  title={Transolver: A fast transformer solver for pdes on general geometries},
  author={Wu, Haixu and Luo, Huakun and Wang, Haowen and Wang, Jianmin and Long, Mingsheng},
  journal={arXiv preprint arXiv:2402.02366},
  year={2024}
}

@article{ranade2025domino,
  title={Domino: A decomposable multi-scale iterative neural operator for modeling large scale engineering simulations},
  author={Ranade, Rishikesh and Nabian, Mohammad Amin and Tangsali, Kaustubh and Kamenev, Alexey and Hennigh, Oliver and Cherukuri, Ram and Choudhry, Sanjay},
  journal={arXiv preprint arXiv:2501.13350},
  year={2025}
}

@article{alkin2025ab,
  title={AB-UPT: Scaling Neural CFD Surrogates for High-Fidelity Automotive Aerodynamics Simulations via Anchored-Branched Universal Physics Transformers},
  author={Alkin, Benedikt and Bleeker, Maurits and Kurle, Richard and Kronlachner, Tobias and Sonnleitner, Reinhard and Dorfer, Matthias and Brandstetter, Johannes},
  journal={arXiv preprint arXiv:2502.09692},
  year={2025}
}

@article{nasacrm,
  title={NASA Common Research Model Project. High speed common research model (crm)},
  author={},
  journal={https://commonresearchmodel.larc.nasa.gov/home-2/high-speed-crm/.},
  year={}
}

@article{kovachki2023neural,
  title={Neural operator: Learning maps between function spaces with applications to pdes},
  author={Kovachki, Nikola and Li, Zongyi and Liu, Burigede and Azizzadenesheli, Kamyar and Bhattacharya, Kaushik and Stuart, Andrew and Anandkumar, Anima},
  journal={Journal of Machine Learning Research},
  volume={24},
  number={89},
  pages={1--97},
  year={2023}
}

@article{lu2021learning,
  title={Learning nonlinear operators via DeepONet based on the universal approximation theorem of operators},
  author={Lu, Lu and Jin, Pengzhan and Pang, Guofei and Zhang, Zhongqiang and Karniadakis, George Em},
  journal={Nature machine intelligence},
  volume={3},
  number={3},
  pages={218--229},
  year={2021},
  publisher={Nature Publishing Group UK London}
}

@article{li2020fourier,
  title={Fourier neural operator for parametric partial differential equations},
  author={Li, Zongyi and Kovachki, Nikola and Azizzadenesheli, Kamyar and Liu, Burigede and Bhattacharya, Kaushik and Stuart, Andrew and Anandkumar, Anima},
  journal={arXiv preprint arXiv:2010.08895},
  year={2020}
}

@article{li2020neural,
  title={Neural operator: Graph kernel network for partial differential equations},
  author={Li, Zongyi and Kovachki, Nikola and Azizzadenesheli, Kamyar and Liu, Burigede and Bhattacharya, Kaushik and Stuart, Andrew and Anandkumar, Anima},
  journal={arXiv preprint arXiv:2003.03485},
  year={2020}
}

@article{li2020multipole,
  title={Multipole graph neural operator for parametric partial differential equations},
  author={Li, Zongyi and Kovachki, Nikola and Azizzadenesheli, Kamyar and Liu, Burigede and Stuart, Andrew and Bhattacharya, Kaushik and Anandkumar, Anima},
  journal={Advances in Neural Information Processing Systems},
  volume={33},
  pages={6755--6766},
  year={2020}
}

@article{li2023fourier,
  title={Fourier neural operator with learned deformations for pdes on general geometries},
  author={Li, Zongyi and Huang, Daniel Zhengyu and Liu, Burigede and Anandkumar, Anima},
  journal={Journal of Machine Learning Research},
  volume={24},
  number={388},
  pages={1--26},
  year={2023}
}

@inproceedings{hao2023gnot,
  title={Gnot: A general neural operator transformer for operator learning},
  author={Hao, Zhongkai and Wang, Zhengyi and Su, Hang and Ying, Chengyang and Dong, Yinpeng and Liu, Songming and Cheng, Ze and Song, Jian and Zhu, Jun},
  booktitle={International Conference on Machine Learning},
  pages={12556--12569},
  year={2023},
  organization={PMLR}
}

@article{bryutkin2024hamlet,
  title={Hamlet: Graph transformer neural operator for partial differential equations},
  author={Bryutkin, Andrey and Huang, Jiahao and Deng, Zhongying and Yang, Guang and Sch{\"o}nlieb, Carola-Bibiane and Aviles-Rivero, Angelica},
  journal={arXiv preprint arXiv:2402.03541},
  year={2024}
}

@article{shih2025transformers,
  title={Transformers as neural operators for solutions of differential equations with finite regularity},
  author={Shih, Benjamin and Peyvan, Ahmad and Zhang, Zhongqiang and Karniadakis, George Em},
  journal={Computer Methods in Applied Mechanics and Engineering},
  volume={434},
  pages={117560},
  year={2025},
  publisher={Elsevier}
}

@article{liu2025geometry,
  title={Geometry-informed neural operator transformer},
  author={Liu, Qibang and Zhong, Weiheng and Meidani, Hadi and Abueidda, Diab and Koric, Seid and Geubelle, Philippe},
  journal={arXiv preprint arXiv:2504.19452},
  year={2025}
}

@article{ovadia2024vito,
  title={Vito: Vision transformer-operator},
  author={Ovadia, Oded and Kahana, Adar and Stinis, Panos and Turkel, Eli and Givoli, Dan and Karniadakis, George Em},
  journal={Computer Methods in Applied Mechanics and Engineering},
  volume={428},
  pages={117109},
  year={2024},
  publisher={Elsevier}
}

@inproceedings{pfaff2020learning,
  title={Learning mesh-based simulation with graph networks},
  author={Pfaff, Tobias and Fortunato, Meire and Sanchez-Gonzalez, Alvaro and Battaglia, Peter},
  booktitle={International conference on learning representations},
  year={2020}
}

@article{pelissier2024graph,
  title={Graph neural networks for mesh generation and adaptation in structural and fluid mechanics},
  author={Pelissier, Ugo and Parret-Fr{\'e}aud, Augustin and Bordeu, Felipe and Mesri, Youssef},
  journal={Mathematics},
  volume={12},
  number={18},
  pages={2933},
  year={2024},
  publisher={MDPI}
}

@article{pilva2022learning,
  title={Learning time-dependent PDE solver using message passing graph neural networks},
  author={Pilva, Pourya and Zareei, Ahmad},
  journal={arXiv preprint arXiv:2204.07651},
  year={2022}
}

@article{gladstone2024mesh,
  title={Mesh-based GNN surrogates for time-independent PDEs},
  author={Gladstone, Rini Jasmine and Rahmani, Helia and Suryakumar, Vishvas and Meidani, Hadi and D’Elia, Marta and Zareei, Ahmad},
  journal={Scientific reports},
  volume={14},
  number={1},
  pages={3394},
  year={2024},
  publisher={Nature Publishing Group UK London}
}

@article{nabian2024x,
  title={X-meshgraphnet: Scalable multi-scale graph neural networks for physics simulation},
  author={Nabian, Mohammad Amin and Liu, Chang and Ranade, Rishikesh and Choudhry, Sanjay},
  journal={arXiv preprint arXiv:2411.17164},
  year={2024}
}

@article{fortunato2022multiscale,
  title={Multiscale meshgraphnets},
  author={Fortunato, Meire and Pfaff, Tobias and Wirnsberger, Peter and Pritzel, Alexander and Battaglia, Peter},
  journal={arXiv preprint arXiv:2210.00612},
  year={2022}
}

@article{choy2025factorized,
  title={Factorized implicit global convolution for automotive computational fluid dynamics prediction},
  author={Choy, Chris and Kamenev, Alexey and Kossaifi, Jean and Rietmann, Max and Kautz, Jan and Azizzadenesheli, Kamyar},
  journal={arXiv preprint arXiv:2502.04317},
  year={2025}
}

@article{chen2025tripnet,
  title={TripNet: Learning Large-scale High-fidelity 3D Car Aerodynamics with Triplane Networks},
  author={Chen, Qian and Elrefaie, Mohamed and Dai, Angela and Ahmed, Faez},
  journal={arXiv preprint arXiv:2503.17400},
  year={2025}
}

@article{eliasof2021pde,
  title={Pde-gcn: Novel architectures for graph neural networks motivated by partial differential equations},
  author={Eliasof, Moshe and Haber, Eldad and Treister, Eran},
  journal={Advances in neural information processing systems},
  volume={34},
  pages={3836--3849},
  year={2021}
}

@inproceedings{liu2021multi,
  title={Multi-resolution graph neural networks for pde approximation},
  author={Liu, Wenzhuo and Yagoubi, Mouadh and Schoenauer, Marc},
  booktitle={International Conference on Artificial Neural Networks},
  pages={151--163},
  year={2021},
  organization={Springer}
}

@article{ranade2022composable,
  title={A composable machine-learning approach for steady-state simulations on high-resolution grids},
  author={Ranade, Rishikesh and Hill, Chris and Ghule, Lalit and Pathak, Jay},
  journal={Advances in Neural Information Processing Systems},
  volume={35},
  pages={17386--17401},
  year={2022}
}

@article{guibas2021adaptive,
  title={Adaptive fourier neural operators: Efficient token mixers for transformers},
  author={Guibas, John and Mardani, Morteza and Li, Zongyi and Tao, Andrew and Anandkumar, Anima and Catanzaro, Bryan},
  journal={arXiv preprint arXiv:2111.13587},
  year={2021}
}

@article{liu2023multiresolution,
  title={Multiresolution convolutional autoencoders},
  author={Liu, Yuying and Ponce, Colin and Brunton, Steven L and Kutz, J Nathan},
  journal={Journal of Computational Physics},
  volume={474},
  pages={111801},
  year={2023},
  publisher={Elsevier}
}

@article{raissi2019physics,
    title={Physics-informed neural networks: A deep learning framework for solving forward and inverse problems involving nonlinear partial differential equations},
    author={Raissi, Maziar and Perdikaris, Paris and Karniadakis, George E},
    journal={Journal of Computational Physics},
    volume={378},
    pages={686--707},
    year={2019},
    publisher={Elsevier}
}

@article{li2024physics,
  title={Physics-informed neural operator for learning partial differential equations},
  author={Li, Zongyi and Zheng, Hongkai and Kovachki, Nikola and Jin, David and Chen, Haoxuan and Liu, Burigede and Azizzadenesheli, Kamyar and Anandkumar, Anima},
  journal={ACM/IMS Journal of Data Science},
  volume={1},
  number={3},
  pages={1--27},
  year={2024},
  publisher={ACM New York, NY}
}

@article{si2025adamuon,
  title={Adamuon: Adaptive muon optimizer},
  author={Si, Chongjie and Zhang, Debing and Shen, Wei},
  journal={arXiv preprint arXiv:2507.11005},
  year={2025}
}

@misc{nvidia2025physicsnemo,
    author = {{NVIDIA}},
    title = {{NVIDIA PhysicsNeMo}},
    year = {2025},
    publisher = {NVIDIA Developer},
    url = {https://developer.nvidia.com/physicsnemo}
}

@article{wen2025goat,
  title        = {Geometry Aware Operator Transformer as an Efficient and Accurate Neural Surrogate for PDEs on Arbitrary Domains},
  author       = {Wen, Shizheng and Kumbhat, Arsh and Lingsch, Levi and Mousavi, Sepehr and Zhao, Yizhou and Chandrashekar, Praveen and Mishra, Siddhartha},
  year         = {2025},
  eprint       = {2505.18781},
  archivePrefix= {arXiv},
  primaryClass = {cs.LG}
}

@inproceedings{satorras2021n,
  title={E (n) equivariant graph neural networks},
  author={Satorras, V{\i}ctor Garcia and Hoogeboom, Emiel and Welling, Max},
  booktitle={International conference on machine learning},
  pages={9323--9332},
  year={2021},
  organization={PMLR}
}

@article{brandstetter2021geometric,
  title={Geometric and physical quantities improve e (3) equivariant message passing},
  author={Brandstetter, Johannes and Hesselink, Rob and van der Pol, Elise and Bekkers, Erik J and Welling, Max},
  journal={arXiv preprint arXiv:2110.02905},
  year={2021}
}

@article{liao2022equiformer,
  title={Equiformer: Equivariant graph attention transformer for 3d atomistic graphs},
  author={Liao, Yi-Lun and Smidt, Tess},
  journal={arXiv preprint arXiv:2206.11990},
  year={2022}
}

@inproceedings{ronneberger2015u,
  title={U-net: Convolutional networks for biomedical image segmentation},
  author={Ronneberger, Olaf and Fischer, Philipp and Brox, Thomas},
  booktitle={International Conference on Medical image computing and computer-assisted intervention},
  pages={234--241},
  year={2015},
  organization={Springer}
}

@article{puri2025flare,
  title={FLARE: Fast Low-rank Attention Routing Engine},
  author={Puri, Vedant and Joglekar, Aditya and Bandreddi, Sri Datta Ganesh and Ferguson, Kevin and Chen, Yu-hsuan and Zhang, Yongjie Jessica and Kara, Levent Burak},
  journal={arXiv preprint arXiv:2508.12594},
  year={2025}
}
\bibliographystyle{unsrtnat}
}

\newpage
\appendix
\section{Model configuration and training details}
\subsection{Datasets}

The benchmarking, validation and ablation studies are carried out on 3 datasets, DrivAerML \cite{ashton2024drivaerml}, Luminary SHIFT-SUV \cite{luminaryshiftsuv} and Luminary SHIFT-Wing \cite{luminaryshiftwing}. Below we provide details about each of these datasets with information such as training, validation and testing splits.

\subsubsection{DrivAerML}
DrivAerML \cite{ashton2024drivaerml} is a public, high-fidelity dataset tailored for AI-based surrogates in automotive external aerodynamics. It comprises 500 parametrically morphed variants of the DrivAer Notchback, providing broad geometric diversity for studying drag and flow behavior at scale. Each case is simulated with a scale-resolving hybrid RANS/LES approach representative of industry practice \cite{spalart2006new, chaouat2017state, heinz2020review, ashton2022hlpw}, on extremely large meshes—on the order of 140 to 150 million volume elements and roughly 9 to 10 million surface points/elements \cite{hupertz2022towards, ashton2024drivaerml}.

For every geometry, time-averaged fields are released in VTK formats: VTP files contain surface quantities (pressure and wall shear stress), and VTU files contain volume fields (velocity, pressure, and turbulence-related variables such as vorticity or turbulent viscosity). The geometry is exported as a coarse STL with around 0.3 million points. The dataset was created to address the lack of open-source, large-scale CFD data suitable for high-fidelity ML research in automotive aerodynamics.

In our study, we adopt a drag-aware split strategy. 10 \% of the samples are held out for testing, and approximately 20\% of this test set is designated out-of-distribution (OOD) based on drag ranges. These OOD cases correspond to some of the lowest and highest drag configurations and are not exposed during training, enabling assessment of generalization to extreme geometries and regimes. Details related to the exact training and validation split may be found in PhysicsNeMo \cite{nvidia2025physicsnemo} here: \url{https://github.com/NVIDIA/physicsnemo-cfd/tree/main/workflows/bench_example/drivaer_ml_files}.

\subsubsection{Luminary SHIFT-SUV}

SHIFT-SUV is an open-source, high-fidelity database of external aerodynamics developed by Luminary Cloud in collaboration with Honda, comprising thousands of transient simulations of parametrically morphed variants of the AeroSUV platform from FKFS (Forschungsinstitut für Kraftfahrwesen und Fahrzeugmotoren Stuttgart) \cite{zhang2019introduction}. AeroSUV shares design lineage and geometric characteristics with the widely used DrivAer platform in sedan aerodynamics \cite{heft2012experimental}. Distributed under a CC-BY-NC license and hosted on HuggingFace \cite{luminaryshiftsuv}, SHIFT-SUV is purpose-built for high-fidelity aerodynamic inference without requiring CFD or meshing expertise, supporting both surface- and volume-based surrogate training, real-time inference, and exploration of shape–performance correlations.

Geometry generation uses a deformation-cage approach implemented in ANSA (BETA-CAE Systems). Cage vertex motions are structured to emulate vehicle design parameters familiar to stylists and defined with guidance from Honda; the parameters are non-orthogonal (multiple parameters may affect shared control points). Configuration options—such as body style (Estate vs. Fastback) and underbody detail (smooth vs. detailed)—are treated as non-parametric toggles. Regions that should not vary across designs (e.g., wheels, tires, suspension) are excluded from morphing to preserve consistency. Specific parameter values are selected via Latin hypercube sampling. The initial dataset varies geometry while keeping boundary conditions fixed. The deformation space uses a morphing‑cage workflow to apply stylist relevant surface translations relative to the baseline model, with ranges specified as target displacements in millimeters that guide, but do not strictly enforce, the final geometry. 

The geometric parameters are varied as follows. Because the morphing cage couples nearby regions, these bounds serve as design targets for the cage control points rather than exact, per‑point displacement limits on the surface. Additional details may be found in the dataset provided by Luminary et al.~\cite{luminaryshiftsuv}.

\begin{itemize}
    \item Front end changes include hood height ($-$50 to $+$50 mm), front (FR) overhang ($-$150 to $+$150 mm), windshield angle implemented via forward/backward shifts ($-$150 to $+$100 mm), and planview adjustments near the front ($-$75 to $+$75 mm),
    \item Global and chassis‑related shifts cover overall vehicle height ($-$150 to $+$150 mm) and ride height ($-$30 to $+$30 mm),
    \item Rear end modifications include backlight angle via surface translation ($-$100 to $+$200 mm), rear (RR) overhang ($-$150 to $+$100 mm), and tapering at both the windshield and rear ($-$100 to $+$100 mm and $-$90 to $+$70 mm, respectively).
\end{itemize}

Variants are organized into four groups by scale and body style—full-scale Estate, full-scale Fastback, quarter-scale Estate, and quarter-scale Fastback—with 998 samples in each group. Quarter-scale cases validate the setup against FKFS wind-tunnel experiments for a 1/4-scale model \cite{zhang2021aerodynamic}, while full-scale cases follow the same methodology with refinements informed by best practices from AutoCFD \cite{economon2024}.

Each sample has a unique volume mesh created in ANSA, yielding hex-dominant and polyhedral meshes. A grid refinement study for the validation configuration selected settings of approximately 45 million cells as a balance between accuracy and cost. All cases were run on the Luminary Cloud platform using a GPU-native, second-order (space and time) finite-volume solver \cite{krakos2025gpu, economon2024}. The turbulence model is transient delayed detached eddy simulation (DDES)—a scale-resolving DES variant—with a shear-layer–adapted length scale and a vortex tilting measure (VTM) to mitigate grey-area effects. The Spalart–Allmaras (SA) model is employed in RANS regions, and a hybrid centered/upwind convective scheme with proprietary blending limits dissipation in LES regions. An advanced shielding function prevents modeled stress depletion, avoiding premature separation. Boundary conditions include a rolling road (translating floor) and rotating wheels, with all other vehicle surfaces treated as no-slip walls. For the full-scale datasets—used in our training—the inflow is a uniform 30 m/s \cite{krakos2025gpu, economon2024}.

For training we use 1996 simulations from the full-scale dataset split randomly into 80/10/10. We match the exact simulations in each split with the AB-UPT benchmarking work in Alkin et al. \cite{alkin2025app}. 

\subsubsection{Luminary SHIFT-Wing}

SHIFT-Wing is an open-source database centered on the NASA Common Research Model (CRM) \cite{nasacrm} for high-speed transonic transport aerodynamics \cite{luminaryshiftwing}. The CRM spans a wide range of configurations—from cruise to high-lift with deployed flaps/slats, optional nacelles/pylons, and empennage—and has been extensively studied experimentally and through community CFD efforts (e.g., AIAA Drag Prediction and High-Lift workshops). Developed in collaboration with Otto Aviation, the dataset is purpose-built for high-fidelity aerodynamic inference of non-linear flow fields without requiring CFD or meshing expertise, supporting both surface- and volume-based surrogate training, real-time inference, and exploration of shape–performance trade-offs.

The current release focuses on the high-speed cruise configuration with only fuselage and wing, emphasizing planform design. A fully parametric CRM was constructed in OnShape by importing the NASA reference, deconstructing it, and reassembling it with exposed variables. The parameterization separates wing and fuselage and proceeds as follows: 

\begin{itemize}
    \item Wing: intersect the reference wing at six spanwise stations to extract airfoil profiles; expose parameters controlling profile translation and rotation (twist); re-loft the wing and reconstruct the tip geometry.
    \item Fuselage: parameterize fuselage length and radius; size the wing–body fairing from the local chord of the newly lofted wing.
    \item Assembly: combine fuselage and wing via Boolean operations. While many micro-parameters are used internally, the dataset varies a compact set of macro planform/fuselage parameters (seven in the current release), with micro-parameters derived from these.
\end{itemize}

Datasets are generated in batches at fixed Mach numbers; geometry parameters and angle of attack are selected via Latin hypercube sampling to span distinct flow regimes intentionally. Lower-Mach batches avoid shocks, whereas higher-Mach batches exhibit complex three-dimensional shock structures, enabling targeted evaluation of AI/ML methods across qualitatively different physics. The SHIFT‑Wing design space targets classical wing planform and fuselage variables around the NASA CRM reference. 

The design and operating parameters are varied as follows. Additional details may be found in the dataset provided by Luminary et al.~\cite{luminaryshiftwing}.
\begin{itemize}
    \item On the fuselage, aspect ratio varies from 7.5 to 11 with a reference value of 9. The quarter‑chord sweep angle spans 25 to 37.5 (reference 35). The root‑chord extension factor ranges from 1.0 to 1.4 (reference 1.373). Fuselage diameter ranges from 240 to 258 inches, with 240 inches as the reference.
    \item On the wing, root twist varies from 3 to 9 (reference 6.717). Spanwise twist is controlled via deltas: root‑to‑break ranges from $-$7 to $-$3 (reference $-$5.953), and break to tip ranges from $-$7.5 to $-$1.5 (reference $-$4.513).
    \item Operating conditions are varied through angle of attack from 0 to 4 and Mach number from 0.5 to 0.85. 
\end{itemize}

All simulations are run on the Luminary Cloud platform using a GPU-native finite-volume solver. For SHIFT-Wing, turbulence is modeled via steady RANS with the Spalart–Allmaras model \cite{krakos2025gpu}. All vehicle surfaces are treated as no-slip walls, and angle of attack is imposed through far-field boundary conditions. A key distinction from SHIFT-SUV is meshing: SHIFT-Wing employs Luminary Mesh Adaptation (LMA), a proprietary solution-adaptive meshing workflow that iteratively refines local mesh density and anisotropy to capture sharp features (e.g., transonic shocks) without user intervention. This automated sequence of meshes and solutions improves accuracy across diverse geometries and flow conditions while eliminating manual meshing effort.

For training, we use 1698 simulations (1138 at Mach 0.5 and 560 at Mach 0.85) from this dataset split randomly into 80/10/10. We match the exact
simulations in each split with the AB-UPT benchmarking work in Alkin et al. \cite{alkin2025app}.

\subsubsection{Crash-dynamics datasets}

As an additional example beyond the aerodynamics datasets, we evaluate GeoTransolver on two nonlinear crash-dynamics benchmarks: a bumper-beam impact dataset and a full-vehicle Body-in-White (BIW) frontal-impact dataset. These datasets test prediction of transient structural response under large deformation, contact, and elastoplastic effects.

The bumper-beam dataset is generated from a parametric finite-element representation of a front-impact bumper-beam scenario. The bumper beam is modeled with shell elements and impacted by a cylindrical rigid wall along the longitudinal direction. The design space varies geometric scaling, impact velocity, shell thickness, and transverse impactor offset, resulting in 135 simulations. The dataset is split into 80\% training, 10\% validation, and 10\% testing.

The BIW crash dataset is generated from a high-fidelity finite-element model of a vehicle Body-in-White undergoing frontal impact against a rigid barrier at 56 km/h. The model contains approximately 400,000 nodes and 380,000 elements. The design space is constructed by perturbing the thicknesses of 33 front-end structural components by $\pm 20\%$ around their nominal values, resulting in 150 simulations. Each simulation spans 120 ms and contains 25 discrete time steps. The dataset is split into 90\% training, 5\% validation, and 5\% testing.

For the BIW benchmark, we additionally evaluate position, velocity, and acceleration histories at driver and passenger toe-pan probe locations, which are sensitive to front-end deformation and are relevant to crashworthiness assessment.

\subsection{Models}

We evaluate GeoTransolver on SHIFT‑SUV and SHIFT‑Wing alongside strong baselines—Transolver and DoMINO—and report AB‑UPT results directly from Alkin et al. \cite{alkin2025app}. To probe architectural sensitivity, we also conduct an ablation study on DrivAerML. Transolver uses the implementation released in PhysicsNeMo, with full self‑attention over concatenated surface and volume tokens (\~10M parameters). DoMINO is run with the default PhysicsNeMo configuration following Ranade et al. \cite{ranade2025domino}, including multi‑scale neighborhoods and auxiliary geometric features (surface normals, areas, signed distance), yielding \~19.7M parameters. AB‑UPT is not re‑implemented or re‑run; we adopt the refined 384‑dimensional results from Alkin et al. \cite{alkin2025app}. GeoTransolver, released in PhysicsNeMo, spans \~10–25M parameters depending on GALE depth, ball‑query radii, and kernel size; Section \ref{ablation} reports accuracy across these architectural choices on DrivAerML. For SHIFT‑Wing, GeoTransolver conditions each block on global parameters (angle of attack, Mach) via geometry/global context projections, in contrast to Transolver’s plain token conditioning. All models are trained for up to 500 epochs on a single NVIDIA GB200 node using the Muon optimizer \cite{si2025adamuon}, under a shared preprocessing and evaluation protocol.

For the crash-dynamics benchmarks, we compare Transolver, GeoTransolver, and GeoTransolver with a FLARE attention backend under matched training protocols. Transolver and GeoTransolver use 128 latent tokens for physics attention, while the FLARE-backend variant uses 128 global queries for low-rank attention. For the bumper-beam dataset, GeoTransolver uses 6 layers, 8 attention heads, 256 hidden channels, and multi-scale ball-query radii of $[0.05, 0.25]$ with $[8, 32]$ neighbors. For the BIW crash dataset, GeoTransolver uses 5 layers, 8 attention heads, and the same multi-scale ball-query configuration. In the FLARE-backend variant, the geometry-aware context pathway is kept fixed and only the self-attention/token-mixing backend is replaced.

\subsection{Metrics}
The prediction accuracy is compared across the metrics mentioned below and averaged over test samples. $j$ refer to the index of the test sample and $\tilde{\cdot}$ denotes the model prediction.

\begin{itemize}
    \item Mean Absolute Error: 
    \begin{equation}
        \frac{1}{N} \sum_{j=1}^{N} \|{u}_j - \tilde{{u}}_j\|
    \end{equation}
    \item Relative $L_1$ Norm: 
    \begin{equation}
        \frac{\sum_{j=1}^{N}\|{u}_j - \tilde{{u}}_j\|}{\sum_{j=1}^{N}\|{u}_j\|}
    \end{equation}
\end{itemize}

For surface predictions, we also compare lift and drag forces on the test geometries. These global metrics are commonly used during design exploration for industrial engineering applications. The total force vector on test geometries is obtained as follows:

\begin{equation}
    {F} = \oint_S (-(p_s - p_{\infty}){\hat{n}} + {\tau}_w)dS,
\end{equation}

where, $p_s$ denotes the surface pressure, ${\tau}_w$ is the shear stress, $\hat{{n}}$ is the normal vector. For calculating the drag force, the normal vector is set in the tangential direction, $n = [1, 0, 0]$, while for lift force it is in the direction perpendicular to flow, $n = [0, 0, 1]$. The lift and drag forces are computed for each of the test cases using both predicted and ground truth fields and the $R^2$ value is calculated over the entire test set, to determine the predictive accuracy. 

For the best-performing GeoTransolver configuration, we provide qualitative assessments: contour plots of key fields (e.g., pressure coefficient, velocity magnitude) and design trends showing predicted drag/lift as functions of representative geometry or regime parameters, alongside ground-truth curves where available.

For crash dynamics, we report the relative $L_2$ error over the predicted spatiotemporal structural response,
\begin{equation}
\epsilon_{L_2}
=
\frac{\sum_j \|\tilde{\mathbf{x}}_j - \mathbf{x}_j\|_2}
{\sum_j \|\mathbf{x}_j\|_2},
\end{equation}
where $\mathbf{x}_j$ and $\tilde{\mathbf{x}}_j$ denote the ground-truth and predicted structural states for test sample $j$. For the BIW benchmark, we also report mean squared error for position, velocity, and acceleration at driver and passenger toe-pan probe locations.

\subsection{Computational cost of multi-scale ball queries}
\label{sec:compute-cost-appendix}

Table~\ref{tab:compute-cost} reports per-step training and inference
throughput for each ball-query configuration on DrivAerML for both
surface and volume tasks (GALE backend, 8$\times$H100, batch size~8,
bfloat16, \texttt{torch.compile}). Mean $\pm$ std are reported across
seeds where available; single-seed entries report mean and std over
within-epoch step times. Single-scale ball queries add 1.1--1.2$\times$
overhead on surface; the 4- and 6-scale multi-radii configurations add
approximately 2$\times$ but plateau between them. The 6-scale entry
required reducing query points from 200k to 160k to fit within H100's
80~GB memory.

\begin{table}[H]
\centering
\caption{Training and inference throughput on DrivAerML
(8$\times$H100, batch size~8, bfloat16,
\texttt{torch.compile}). Mean $\pm$ std reported.}
\label{tab:compute-cost}
\begin{small}
\begin{tabular}{@{}l l cc cc@{}}
\toprule
& & \multicolumn{2}{c}{Surface} & \multicolumn{2}{c}{Volume} \\
\cmidrule(lr){3-4} \cmidrule(lr){5-6}
Config & Radii
  & \makecell{Train\\(s/step)}
  & \makecell{Infer\\(s/step)}
  & \makecell{Train\\(s/step)}
  & \makecell{Infer\\(s/step)} \\
\midrule
No local feat.
  & ---
  & $0.258 \pm 0.002$
  & $0.120 \pm 0.002$
  & $0.436 \pm 0.007$
  & $0.227 \pm 0.008$ \\
1 scale ($r{=}0.05$)
  & {[}16{]}
  & $0.299 \pm 0.003$
  & $0.131 \pm 0.004$
  & $0.487 \pm 0.011$
  & $0.252 \pm 0.009$ \\
1 scale ($r{=}2.5$)
  & {[}64{]}
  & $0.290 \pm 0.001$
  & $0.120 \pm 0.001$
  & $0.611 \pm 0.004$
  & $0.352 \pm 0.011$ \\
4 scales
  & {[}16,32,32,64{]}
  & $0.544 \pm 0.014$
  & $0.254 \pm 0.015$
  & $0.858 \pm 0.022$
  & $0.512 \pm 0.009$ \\
6 scales$^{*}$
  & {[}8,16,32,32,64,128{]}
  & $0.548 \pm 0.010$
  & $0.254 \pm 0.012$
  & $0.794 \pm 0.036$
  & $0.503 \pm 0.010$ \\
\bottomrule
\multicolumn{6}{@{}l}{\footnotesize $^{*}$160k query points (other configs use 200k) to fit in GPU memory.}
\end{tabular}
\end{small}
\end{table}

\subsection{Query and geometry token ablation}

Table~\ref{tab:results-l1-drivaerml-sampling} reports relative $L_1$ errors on DrivAerML as a function of query and geometry token counts. Increasing geometry token density at fixed queries sharply reduces volume errors (e.g., at 20k queries, $p_v$ drops from 11.1\% at 50k geo to 3.09\% at 300k), highlighting the need for adequate geometric coverage. Scaling query tokens improves surface metrics, with the best overall surface/integral performance at 60k/300k. Volume pressure exhibits a sweet spot at 60k/150k, whereas very high geometry counts with moderate queries can degrade volume errors, suggesting imbalance.

\begin{table}[H]
\centering
\caption{Relative $L_1$ errors (\%) on DrivAerML test geometries for number of geometry and query points.}
\label{tab:results-l1-drivaerml-sampling}
\begin{small}
\begin{tabular}{l cccc cc}
\toprule
\cmidrule(lr){2-7}
& \multicolumn{4}{c}{Surface} & \multicolumn{2}{c}{Volume} \\
\cmidrule(lr){2-5} \cmidrule(lr){6-7}
Query/Geo & $p_s$ & $\tau_w$ & $C_D$ & $C_L$ & $p_v$ & $u$ \\
\midrule
 20k/50k &  3.08  & 5.36  & 0.993  & 0.989 & 11.1 & 9.3 \\
 20k/150k & 3.02  & 5.29  & 0.994  & 0.989 & 7.03 & 6.74 \\
 20k/300k & 3.04  & 5.27  & 0.993  & 0.987 & 3.09 & 4.01 \\
 40k/50k & 3.03  & 5.22  & 0.993  & 0.985 & 3.01 & \textbf{3.97} \\
 40k/150k & 2.96 & 5.20  & 0.995  & 0.988 & 3.13 & \textbf{3.97} \\
 40k/300k & 2.98  & 5.25  & 0.994  & 0.987 & 4.41 & 4.76 \\
 60k/50k & 2.97  & 5.17  & 0.994  & 0.987 & 5.61 & 5.75 \\
 60k/150k & 2.99  & 5.15  & 0.994  & 0.987 & \textbf{2.96} & 4.02 \\
 60k/300k & \textbf{2.86}  & \textbf{4.9}  & \textbf{0.996}  & \textbf{0.991} & 3.01 & 4.02 \\
\bottomrule
\end{tabular}
\end{small}
\end{table}

\section{Additional results}

In this section, we provide additional results for the three aerodynamics datasets evaluated in this paper, DrivAerML, SHIFT-SUV, and SHIFT-Wing, as well as the two crash-dynamics benchmarks. For the aerodynamics datasets, we compare GeoTransolver predictions with simulated ground truth to assess its ability to capture design trends and surface and volume contours. For crash dynamics, we assess deformation fields and safety-relevant probe responses.

\subsection{DrivAerML}
\begin{figure}[ht]
    \centering
    \includegraphics[width=0.7\linewidth]{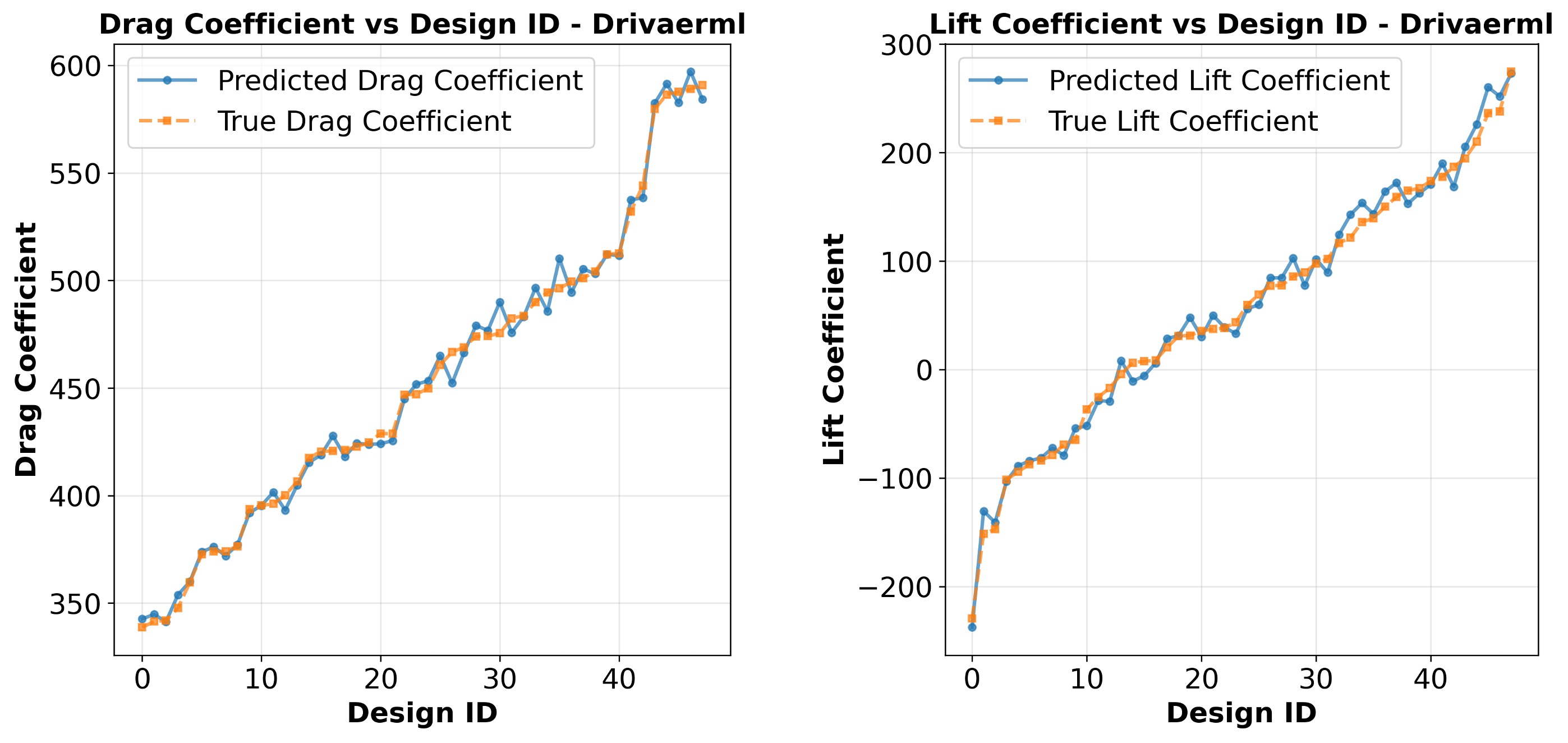}
    \par
    \caption{Design trends: drag and lift vs.\ test designs.}
    \label{fig:drivaerml_desigm}
\end{figure}

Figure \ref{fig:drivaerml_desigm} shows drag and lift coefficients across the 48 test designs, with designs ordered by ascending ground-truth values. The predicted curves closely track the ground-truth trends, preserving the overall ordering and capturing large-scale variations. Small oscillations appear between adjacent designs with subtle directional changes, indicating minor non-monotonic deviations in tightly clustered regions. The design points with the highest and lowest drag and lift forces correspond to out-of-distribution and are never seen during training. It may be noted that the model captures the aerodynamic forces for these designs reasonably accurately and the predictions can be further improved with augmentation of the training set.

\subsection{Surface-field SHIFT results}
\label{sec:shift-surface}

Table~\ref{tab:shift-surface} reports the surface-field counterpart
to the volume table in the main text
(Table~\ref{tab:shift-volume}). GeoTransolver leads on most
surface metrics with two exceptions: surface pressure at Mach 0.85
(AB-UPT) and wall-shear stress at Mach 0.5 (DoMINO, trained
per-Mach).

\begin{table}[H]
\centering
\caption{Surface-field relative $L_1$ errors (\%) across SHIFT-SUV
(Estate, Fastback) and SHIFT-Wing (Ma=0.5, Ma=0.85). DoMINO is
trained per-Mach on SHIFT-Wing; other models share the combined
dataset.}
\label{tab:shift-surface}
\setlength{\tabcolsep}{4pt}
\begin{small}
\begin{tabular}{@{}l cc cc cc cc@{}}
\toprule
& \multicolumn{2}{c}{Estate} & \multicolumn{2}{c}{Fastback}
& \multicolumn{2}{c}{Ma=0.5} & \multicolumn{2}{c}{Ma=0.85} \\
\cmidrule(lr){2-3}\cmidrule(lr){4-5}\cmidrule(lr){6-7}\cmidrule(lr){8-9}
Model & $p_s$ & $\tau_w$ & $p_s$ & $\tau_w$ & $p_s$ & $\tau_w$ & $p_s$ & $\tau_w$ \\
\midrule
GeoTransolver & \textbf{0.0057} & \textbf{3.81} & \textbf{0.0056} & \textbf{3.70}
              & \textbf{0.021}  & 12.2          & 0.081           & \textbf{13.01} \\
AB-UPT        & 0.0064          & 4.95          & 0.0064          & 5.03
              & 0.022           & 12.5          & \textbf{0.079}  & 13.3 \\
DoMINO        & 0.0100          & 12.24         & 0.0100          & 11.74
              & 0.468           & \textbf{10.2} & 1.88            & 13.35 \\
Transolver    & 0.0079          & 4.98          & 0.0078          & 4.97
              & 0.094           & 12.4          & 0.098           & 13.2 \\
\bottomrule
\end{tabular}
\end{small}
\end{table}

\subsection{SHIFT-SUV}
\begin{figure}[ht]
    \centering
    \includegraphics[width=0.7\linewidth]{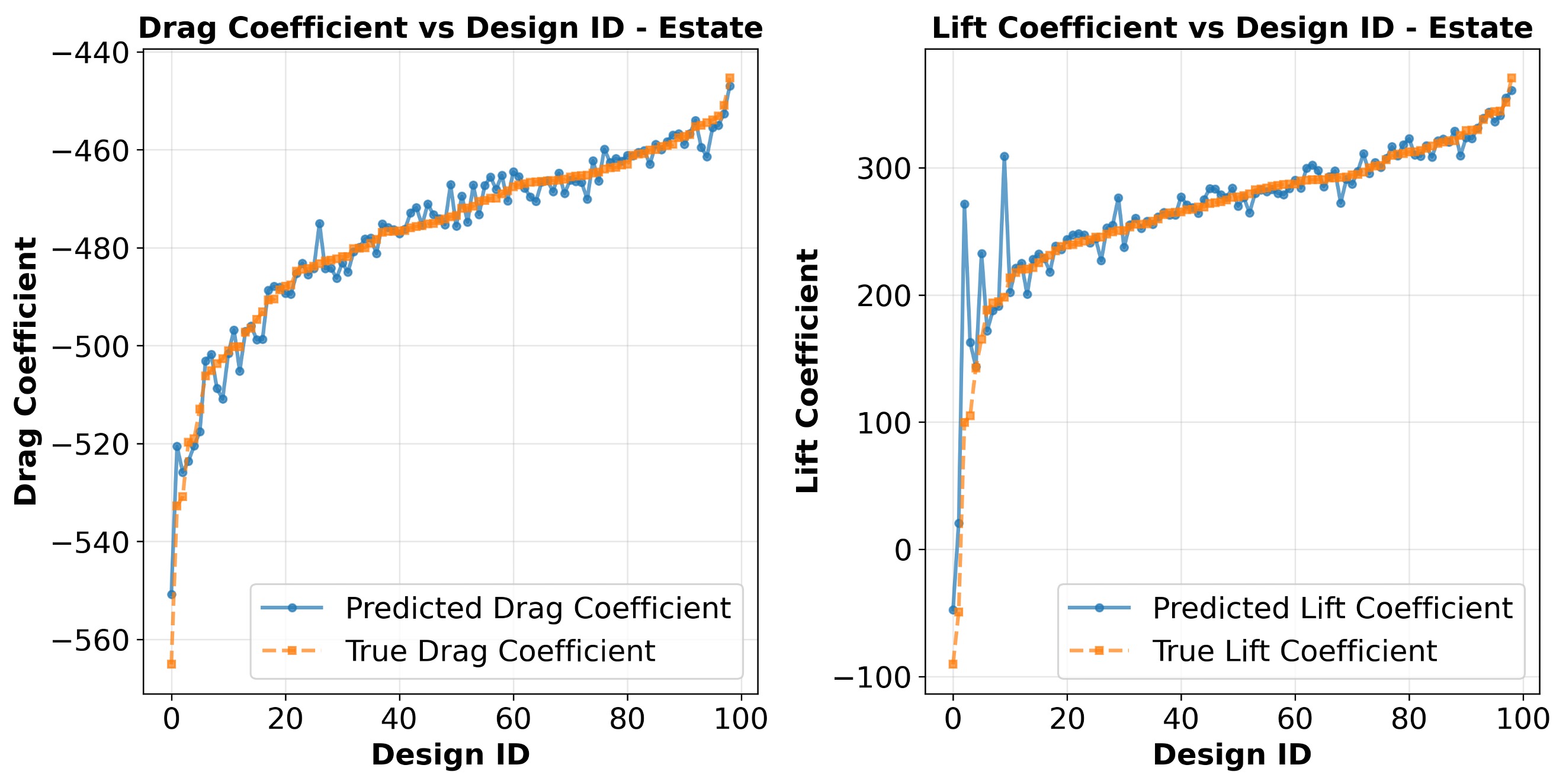}
    \par
    \caption{Design trends for Estate: drag and lift vs.\ test designs.}
    \label{fig:Estate_desigm}
\end{figure}

\begin{figure}[ht]
    \centering
    \includegraphics[width=0.7\linewidth]{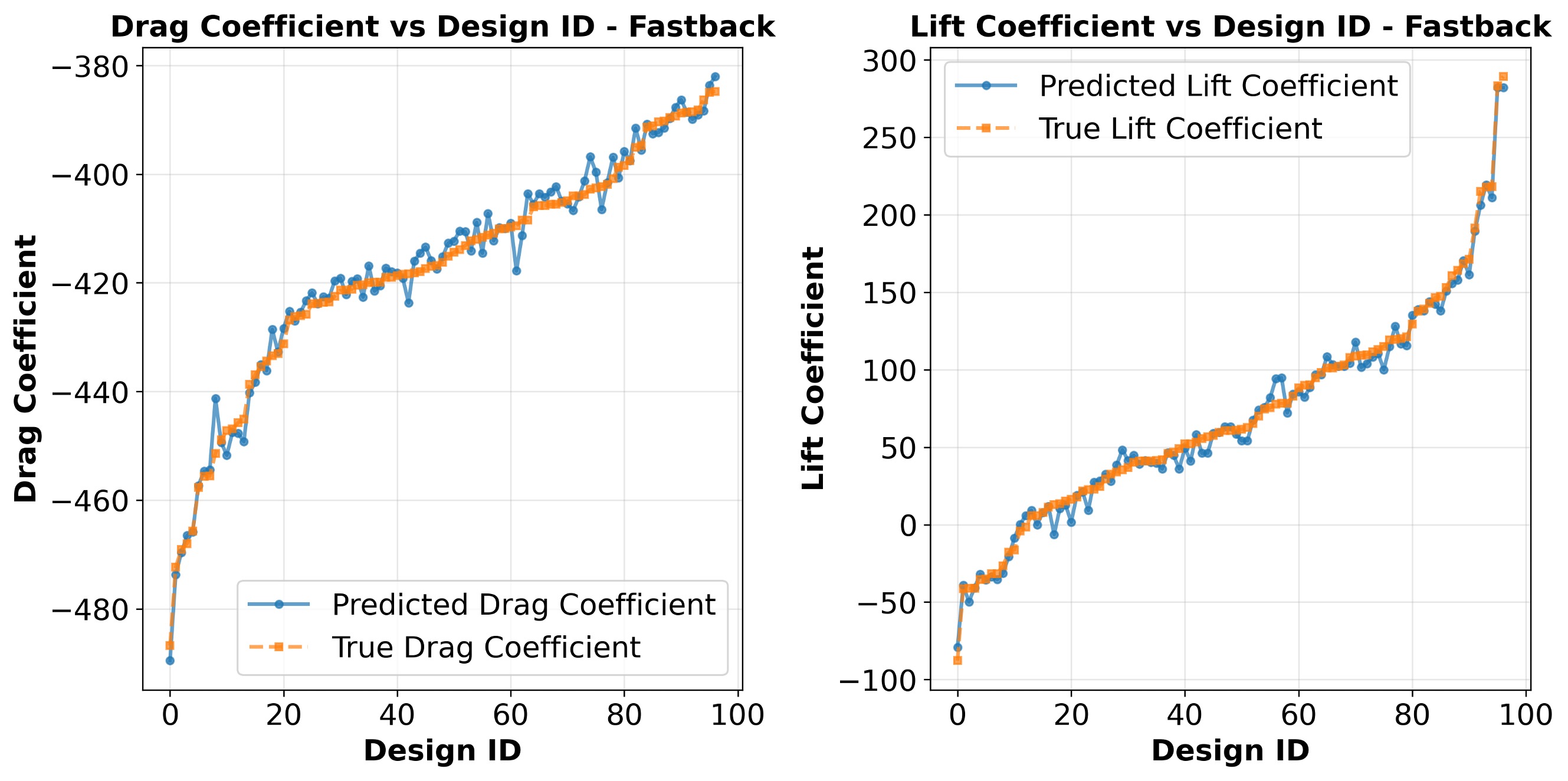}
    \par
    \caption{Design trends for Fastback: drag and lift vs.\ test designs.}
    \label{fig:Fastback_desigm}
\end{figure}

Figures \ref{fig:Estate_desigm} and \ref{fig:Fastback_desigm} plot drag and lift coefficients across the test design IDs for the Estate and Fastback. The predicted trends closely follow ground truth, capturing overall ordering and directional changes. Larger oscillations are visible in Estate lift, likely due to sparser coverage in that lift regime. 

\begin{figure}[ht]
    \centering
    \includegraphics[width=0.7\linewidth]{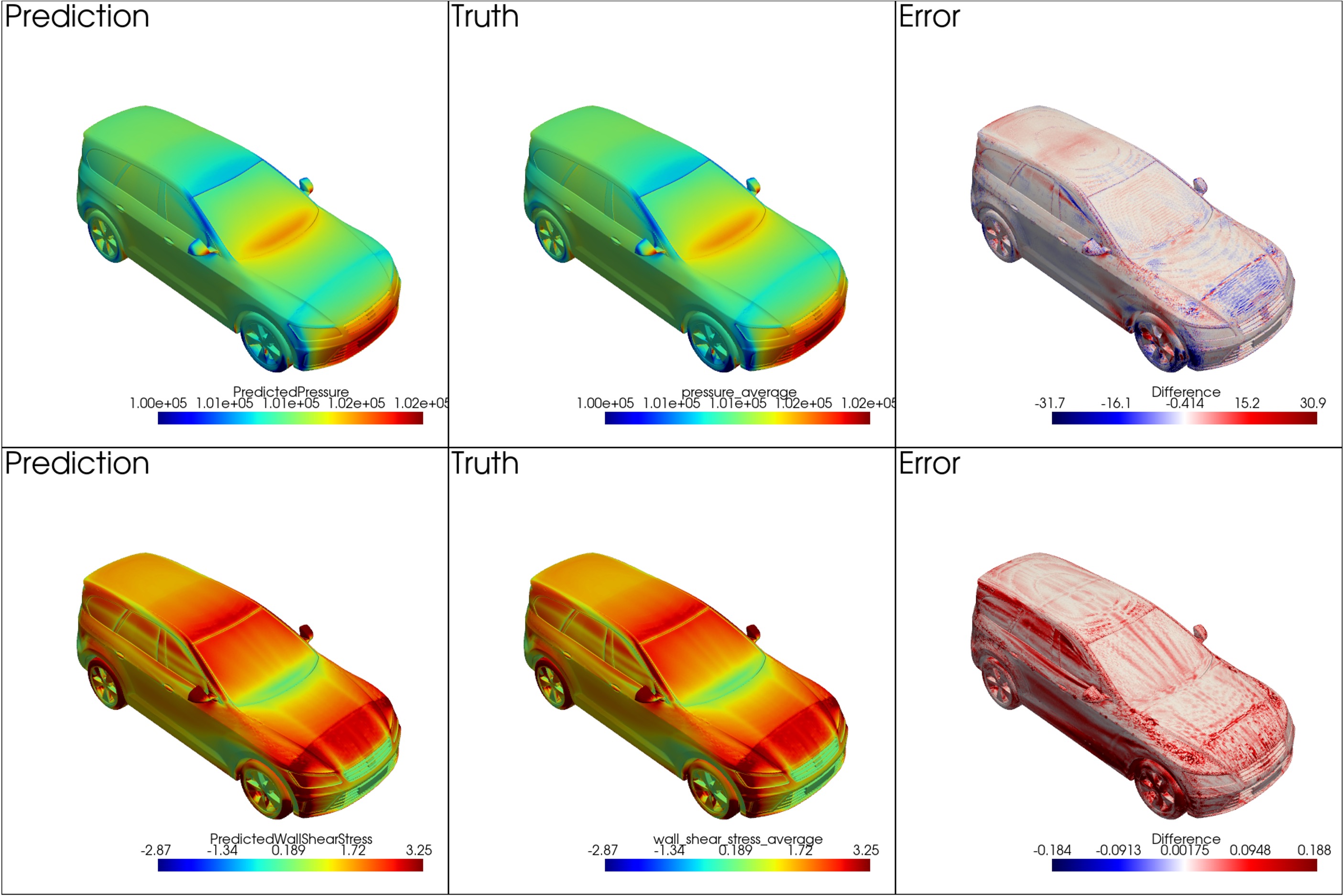}
    \par
    \caption{Surface contour comparisons for Estate.}
    \label{fig:Estate_surface}
\end{figure}

\begin{figure}[ht]
    \centering
    \includegraphics[width=0.7\linewidth]{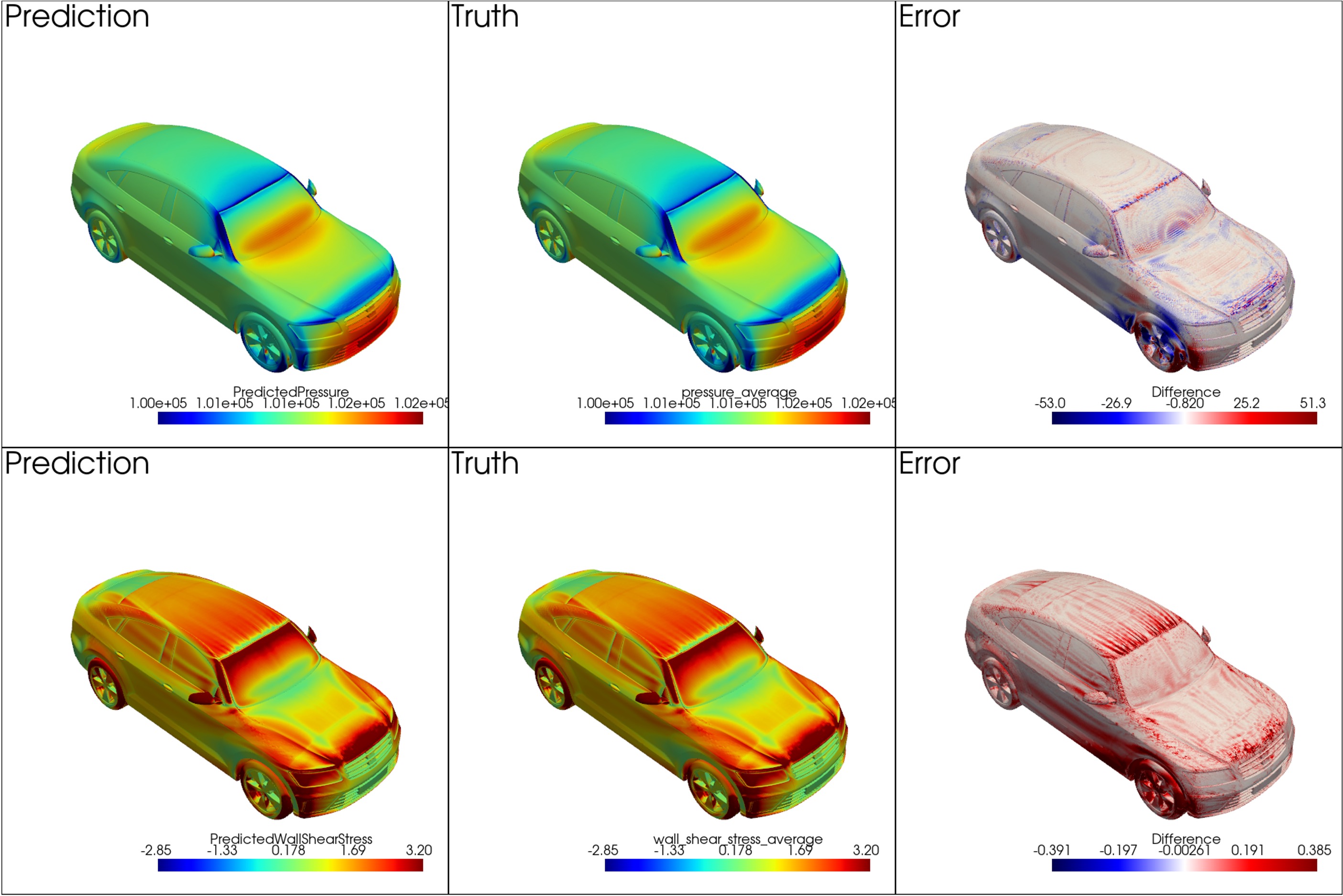}
    \par
    \caption{Surface contour comparisons for Fastback.}
    \label{fig:Fastback_surface}
\end{figure}

Figures \ref{fig:Estate_surface} and \ref{fig:Fastback_surface} compare surface pressure and wall-shear stress contours between GeoTransolver and ground truth. Errors are uniformly distributed and bounded, indicating high accuracy; the largest discrepancies occur near wheels and mirrors, where flow is particularly complex.

\begin{figure}[ht]
    \centering
    \includegraphics[width=0.5\linewidth]{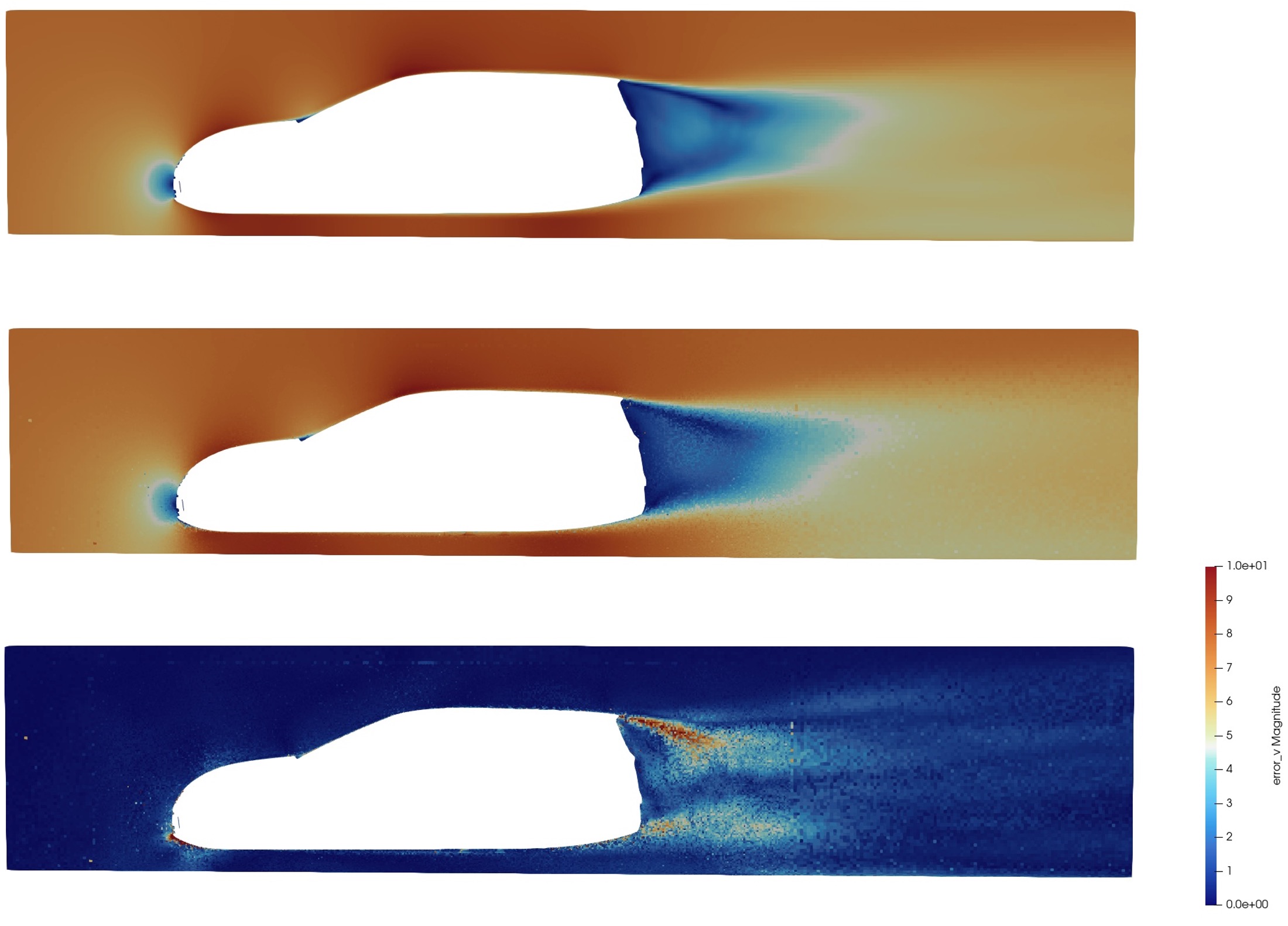}
    \par
    \caption{Volume contour comparisons for Estate on the $xz$ plane.}
    \label{fig:Estate_volume}
\end{figure}

\begin{figure}[ht]
    \centering
    \includegraphics[width=0.5\linewidth]{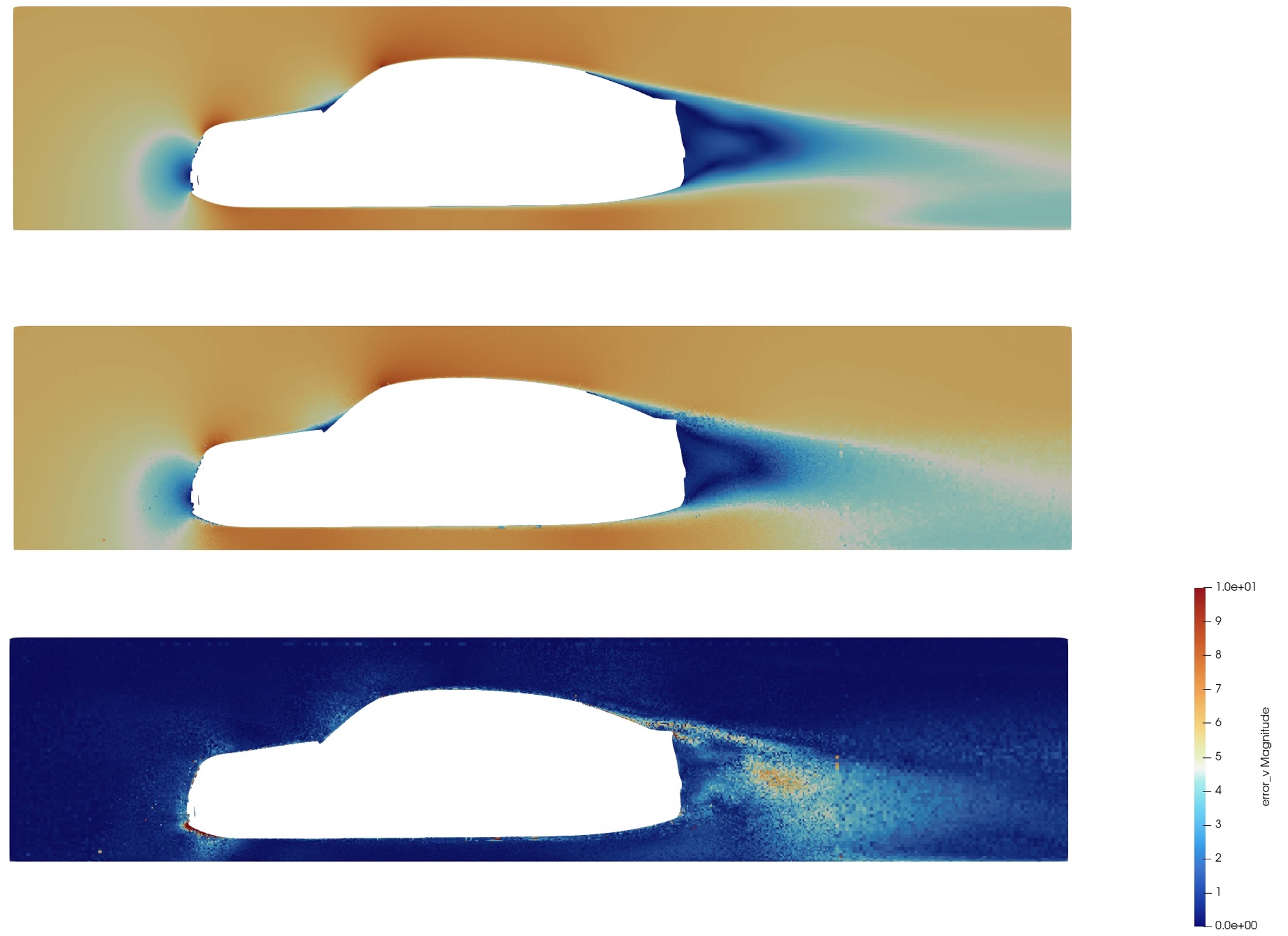}
    \par
    \caption{Volume contour comparisons for Fastback on the $xz$ plane.}
    \label{fig:Fastback_volume}
\end{figure}

Similarly, Figures \ref{fig:Estate_volume} and \ref{fig:Fastback_volume} show volume comparisons on the $xz$ plane. The highest errors appear near separation regions downstream and in the wake where gradients are sharp, but overall predictions remain accurate.


\subsection{SHIFT-Wing}
\begin{figure}[ht]
    \centering
    \includegraphics[width=0.7\linewidth]{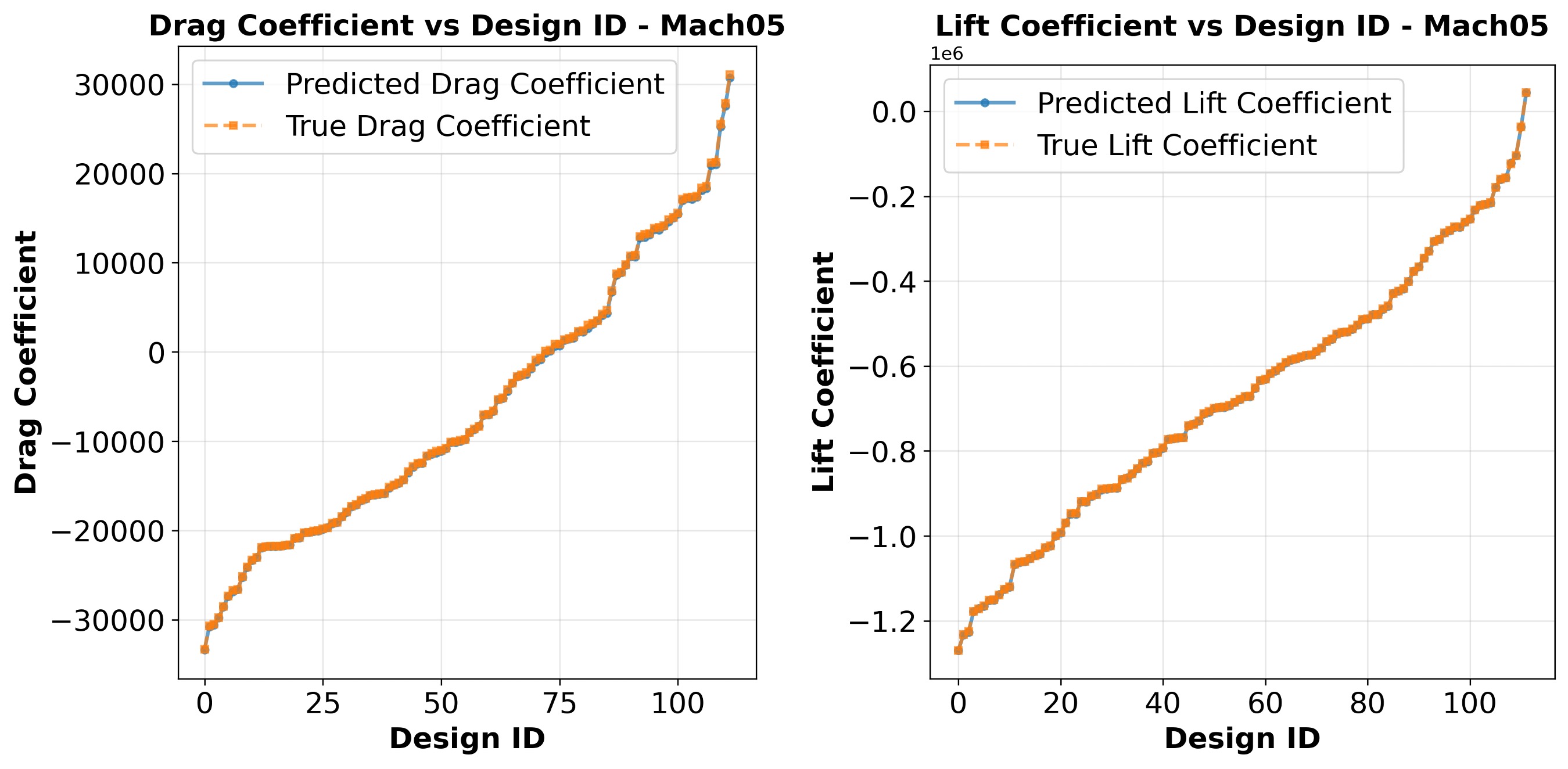}
    \par
    \caption{Design trends for Mach 0.5: drag and lift vs.\ test designs.}
    \label{fig:mach_05_design}
\end{figure}

\begin{figure}[ht]
    \centering
    \includegraphics[width=0.7\linewidth]{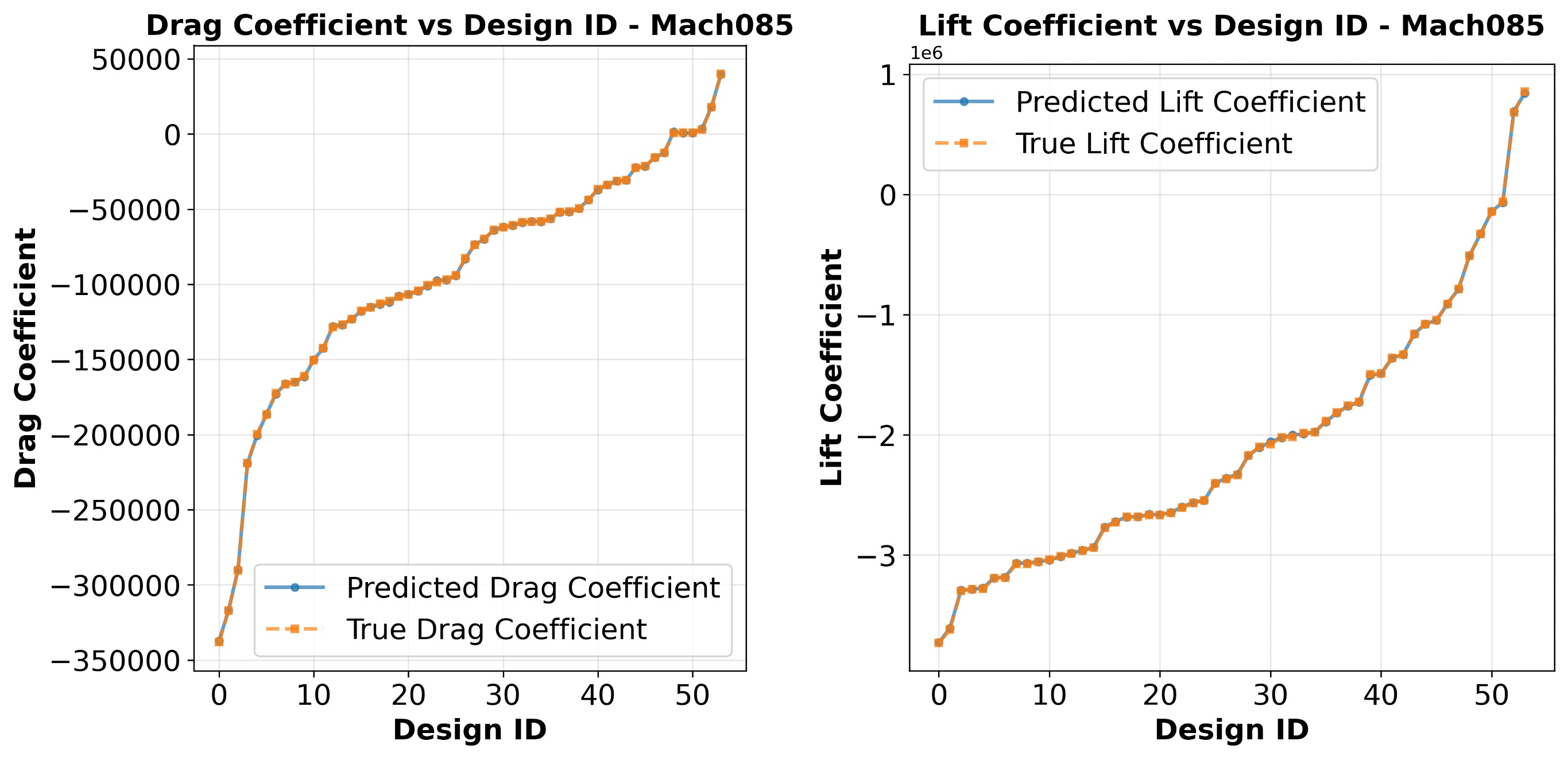}
    \par
    \caption{Design trends for Mach 0.85: drag and lift vs.\ test designs.}
    \label{fig:mach_085_design}
\end{figure}

Figures \ref{fig:mach_05_design} and \ref{fig:mach_085_design} plot drag and lift across test design IDs at Mach 0.5 and 0.85. Predicted trends closely track ground truth in both regimes, preserving ordering and directional changes.

\begin{figure}[ht]
    \centering
    \includegraphics[width=0.7\linewidth]{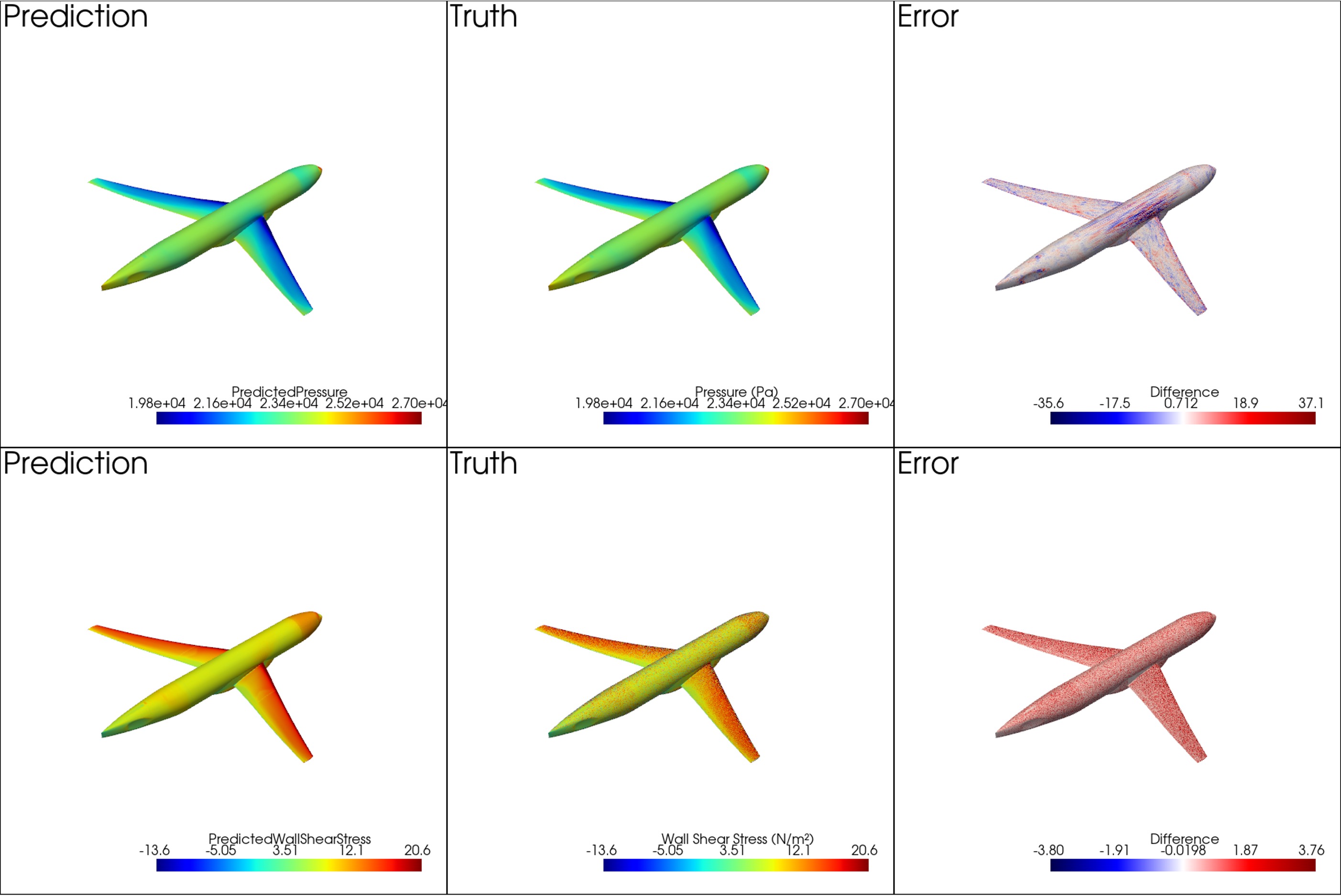}
    \par
    \caption{Surface contour comparisons for Mach 0.5.}
    \label{fig:mach_05_surface}
\end{figure}

\begin{figure}[ht]
    \centering
    \includegraphics[width=0.7\linewidth]{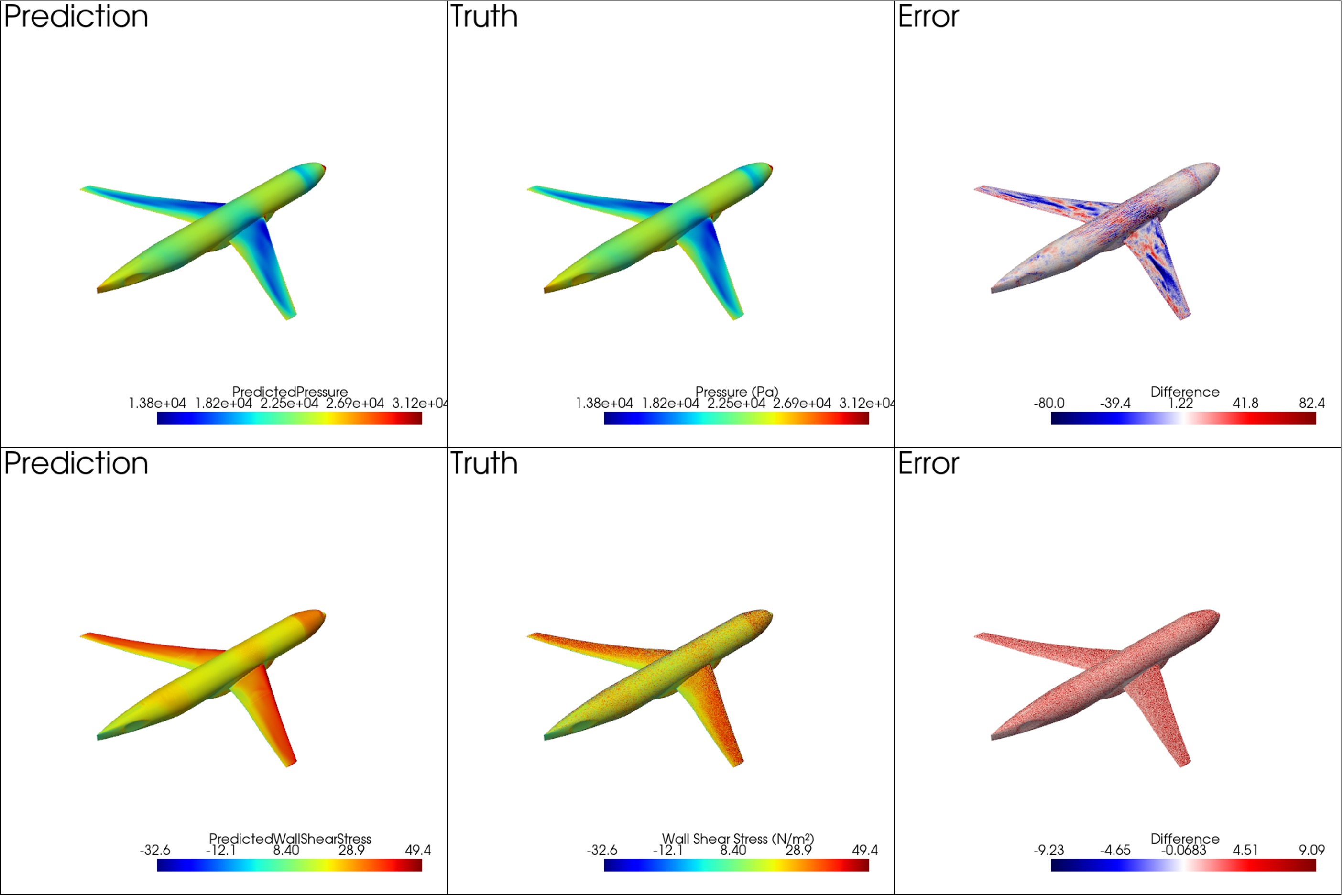}
    \par
    \caption{Surface contour comparisons for Mach 0.85.}
    \label{fig:mach_085_surface}
\end{figure}

Figures \ref{fig:mach_05_surface} and \ref{fig:mach_085_surface} compare surface pressure and wall-shear stress contours between GeoTransolver and ground truth. Errors are uniformly distributed and bounded, indicating high accuracy; the largest discrepancies appear on the wing, especially at higher Mach where sharper transitions occur.

\begin{figure}[ht]
    \centering
    \includegraphics[width=0.9\linewidth]{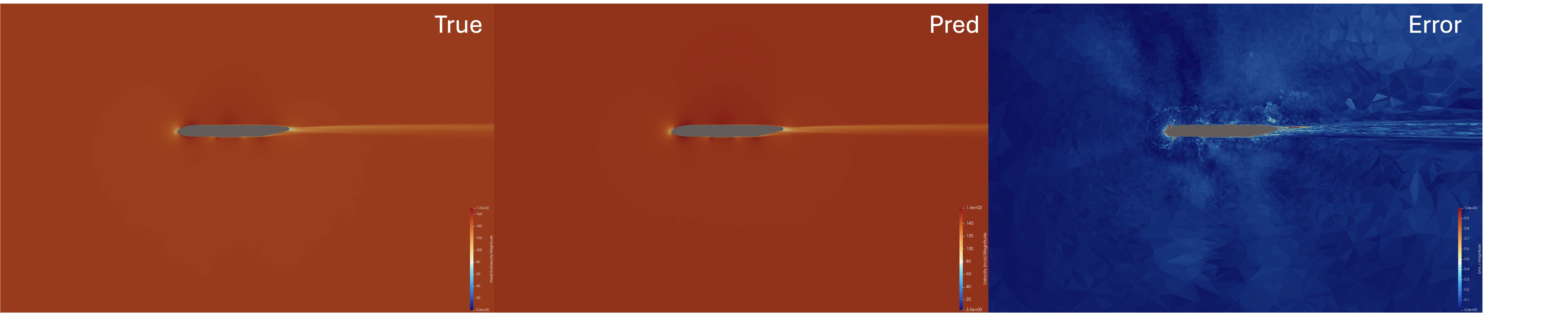}
    \par
    \caption{Volume contour comparisons on the $xz$ plane for Mach 0.5.}
    \label{fig:mach_05_volume}
\end{figure}

\begin{figure}[ht]
    \centering
    \includegraphics[width=0.9\linewidth]{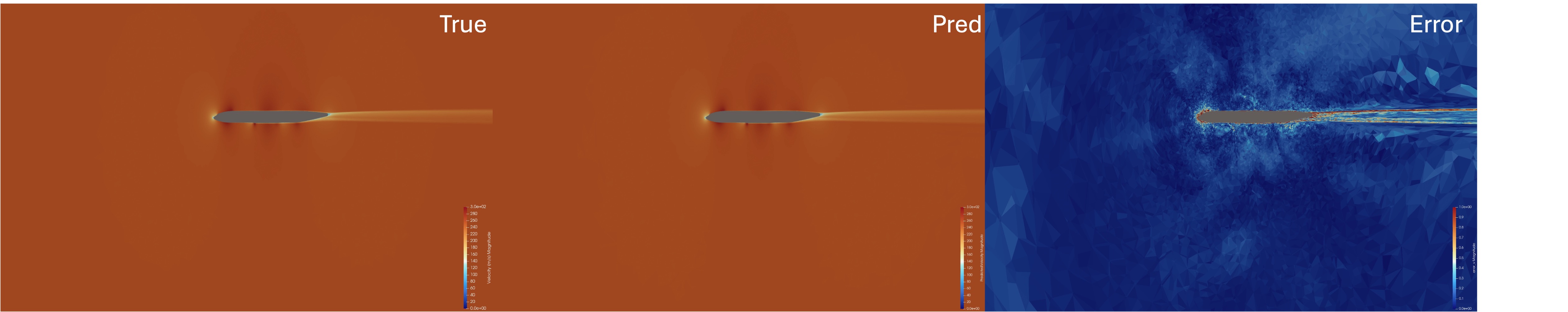}
    \par
    \caption{Volume contour comparisons on the $xz$ plane for Mach 0.85.}
    \label{fig:mach_085_volume}
\end{figure}

Figures \ref{fig:mach_05_volume} and \ref{fig:mach_085_volume} show volume comparisons on the $xz$ plane. Errors are most pronounced over the wing and fuselage and in the downstream wake where separation and steep gradients occur, particularly at Mach 0.85. Overall, predictions remain accurate across both regimes.


\subsection{Crash dynamics}

In this section, we provide additional results for the bumper-beam and BIW crash-dynamics datasets. We compare GeoTransolver predictions with finite-element ground truth to assess its ability to capture final deformation states, deformation evolution over time, and safety-relevant probe responses.

\begin{table}[H]
\centering
\caption{Training speed and number of parameters for the BIW crash-dynamics benchmark using the Muon optimizer.}
\label{tab:crash_speed_params}
\begin{tabular}{lcc}
\toprule
Model & Training speed & No. of parameters \\
\midrule
Transolver & 4.8 s/epoch & 3,700,783 \\
GeoTransolver & 7.5 s/epoch & 5,686,504 \\
GeoTransolver with FLARE & 7.1 s/epoch & 5,150,624 \\
\bottomrule
\end{tabular}
\end{table}

\begin{table}[H]
\centering
\caption{Probe-level MSE comparison for BIW crash dynamics using the Muon optimizer. Position, velocity, and acceleration are evaluated at driver and passenger toe-pan locations.}
\label{tab:crash_probe_mse}
\begin{tabular}{llccc}
\toprule
Probe & Model & Position & Velocity & Acceleration \\
\midrule
Driver & Transolver & $2.63 \times 10^{-3}$ & $8.41 \times 10^{-1}$ & $2.14 \times 10^{3}$ \\
Driver & GeoTransolver & $1.84 \times 10^{-3}$ & $6.09 \times 10^{-1}$ & $1.71 \times 10^{3}$ \\
Driver & GeoTransolver with FLARE & $\mathbf{7.18 \times 10^{-4}}$ & $\mathbf{2.71 \times 10^{-1}}$ & $\mathbf{1.27 \times 10^{3}}$ \\
\midrule
Passenger & Transolver & $2.21 \times 10^{-3}$ & $7.25 \times 10^{-1}$ & $2.24 \times 10^{3}$ \\
Passenger & GeoTransolver & $1.72 \times 10^{-3}$ & $5.53 \times 10^{-1}$ & $1.80 \times 10^{3}$ \\
Passenger & GeoTransolver with FLARE & $\mathbf{6.52 \times 10^{-4}}$ & $\mathbf{2.53 \times 10^{-1}}$ & $\mathbf{1.45 \times 10^{3}}$ \\
\bottomrule
\end{tabular}
\end{table}

Table~\ref{tab:crash_probe_mse} shows that GeoTransolver improves the prediction of position, velocity, and acceleration at both driver and passenger probe locations, and that replacing the physics-attention backend with FLARE further improves these safety-relevant quantities. These probe-level results complement the full-field relative $L_2$ errors reported in the main text and indicate that the geometry-aware context remains effective when paired with an alternative efficient attention backend.

\begin{figure}[ht]
\centering
\includegraphics[width=\linewidth]{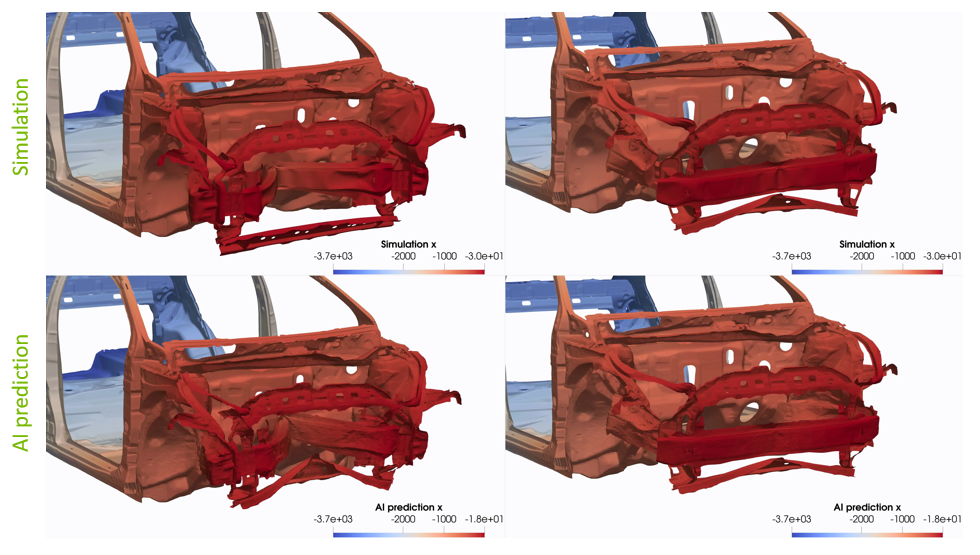}
\caption{Visual comparison of BIW crash deformation for representative test samples. The top row shows finite-element ground truth, and the bottom row shows GeoTransolver predictions using the FLARE attention backend.}
\label{fig:biw_crash_deformation_comparison}
\end{figure}

\begin{figure}[ht]
\centering
\begin{subfigure}{\linewidth}
    \centering
    \includegraphics[width=\linewidth]{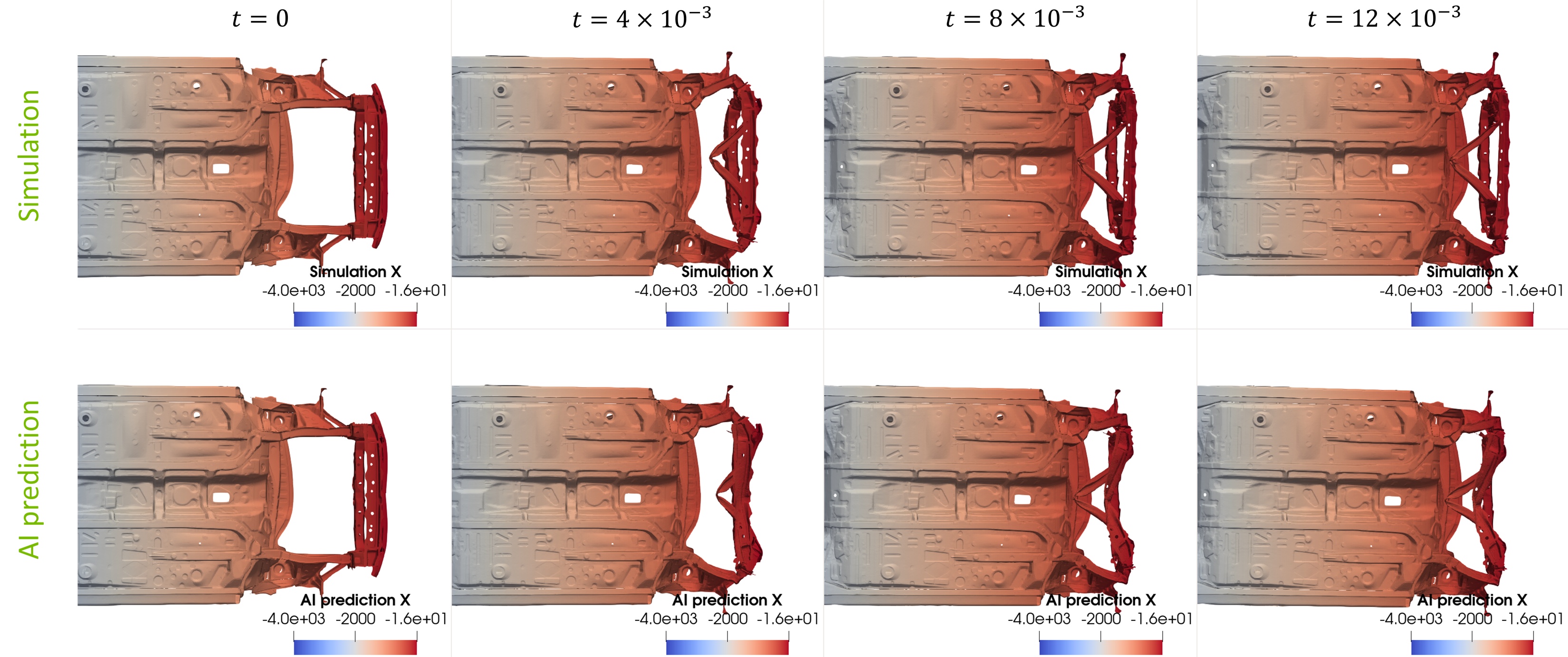}
    \caption{BIW crash deformation evolution for the first held-out test sample.}
    \label{fig:biw_crash_time_evolution_a}
\end{subfigure}
\\[2ex]
\begin{subfigure}{\linewidth}
    \centering
    \includegraphics[width=\linewidth]{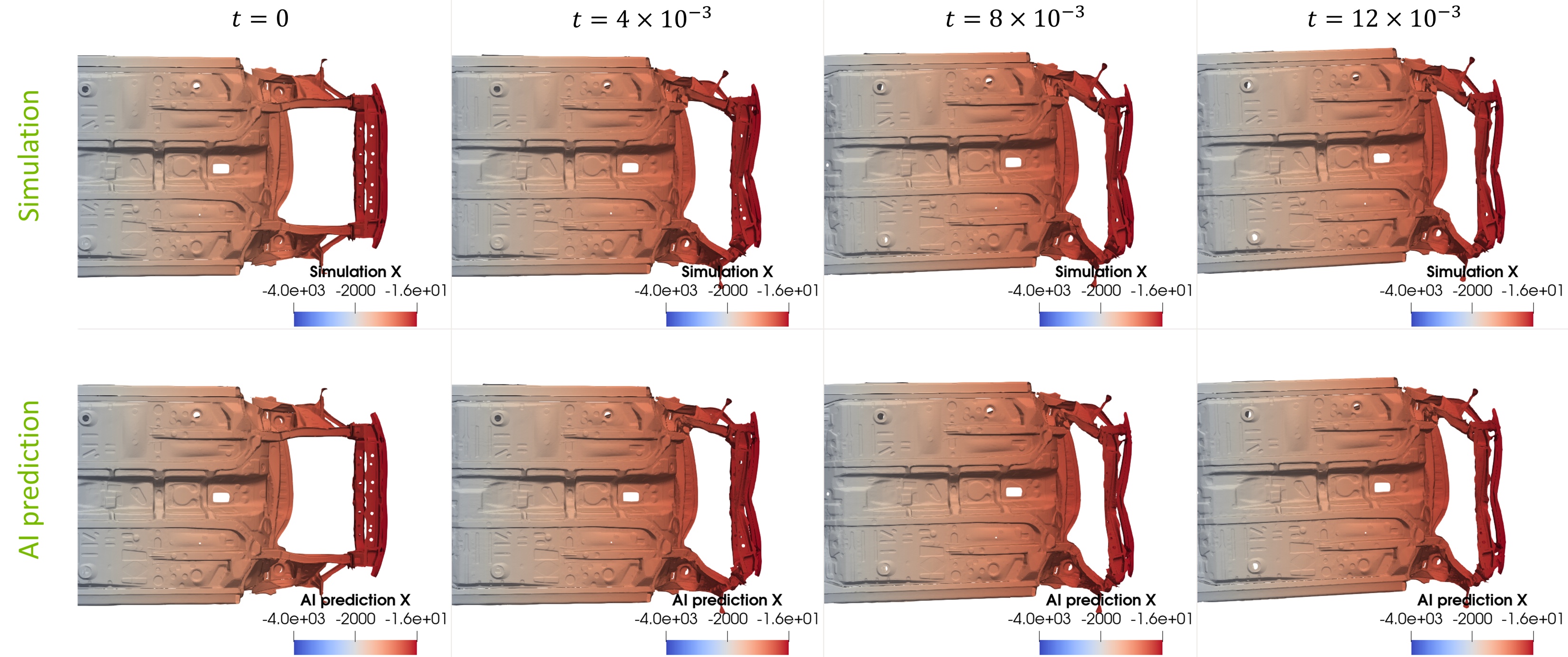}
    \caption{BIW crash deformation evolution for the second held-out test sample.}
    \label{fig:biw_crash_time_evolution_b}
\end{subfigure}
\caption{Temporal evolution of BIW crash deformation for the same held-out test samples shown in Fig.~\ref{fig:biw_crash_deformation_comparison}. Columns show representative time steps through the impact event. For each sample, the top row shows finite-element ground truth and the bottom row shows GeoTransolver predictions using the FLARE attention backend.}
\label{fig:biw_crash_time_evolution}
\end{figure}

\begin{figure}[ht]
\centering
\includegraphics[width=0.75\linewidth]{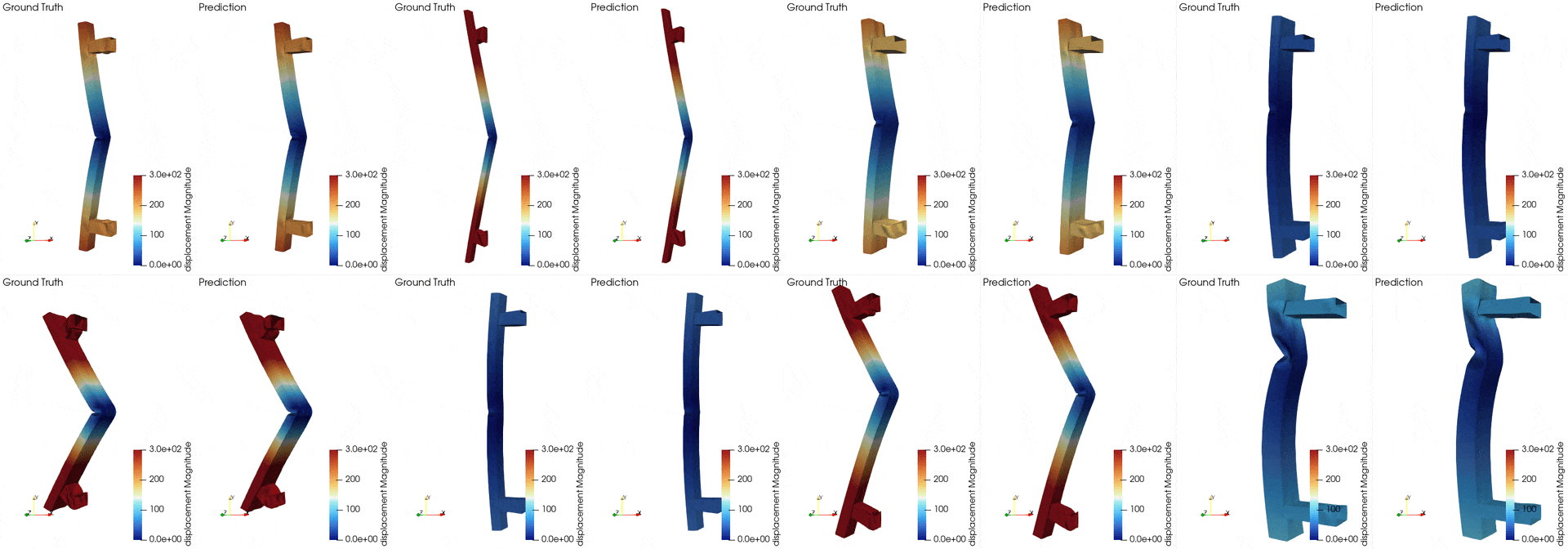}
\caption{Visual comparison of bumper-beam crash deformation for representative test samples. The top row shows finite-element ground truth, and the bottom row shows GeoTransolver predictions using the FLARE attention backend.}
\label{fig:bumper_crash_deformation_comparison}
\end{figure}

Figures~\ref{fig:biw_crash_deformation_comparison}, \ref{fig:biw_crash_time_evolution}, and~\ref{fig:bumper_crash_deformation_comparison} provide qualitative deformation comparisons for GeoTransolver with the FLARE attention backend on the crash-dynamics benchmarks. These visual results help localize prediction differences during large deformation and contact-dominated response.


\clearpage
\FloatBarrier

\end{document}